\documentclass{article}

\usepackage{arxiv}

\usepackage{graphicx}%
\usepackage{amsmath,amssymb,amsfonts}%
\usepackage{amsthm}%
\usepackage{mathrsfs}%
\usepackage[title]{appendix}%
\usepackage{xcolor}%
\usepackage{textcomp}%
\usepackage{manyfoot}%
\usepackage{booktabs}%
\usepackage{listings}%
\usepackage[utf8]{inputenc}
\usepackage{times}
\usepackage{microtype}
\usepackage{amsbsy,amscd}
\usepackage{array}
\usepackage{latexsym}
\usepackage{subcaption}
\usepackage[numbers]{natbib}
\usepackage{colortbl}
\usepackage{multicol}
\usepackage{accents}
\usepackage[exponent-product = \times]{siunitx}
\usepackage{mathtools}
\usepackage{bm}
\usepackage{hyperref}

\newcommand{\D}{\mathrm d}
\DeclareMathOperator{\grad}{grad}
\newcommand{\dif}[2]{\frac{\displaystyle \partial #1}{\displaystyle \partial #2}}
\newcommand{\difn}[2]{\frac{\displaystyle \D #1}{\displaystyle \D #2}}

\newcommand{\Th}[1]{\ensuremath{\hat{\mathrm{\mathbf{#1}}}} }

\newcommand{\LM}[1]{\text{\sffamily\bfseries{#1}}}
\newcommand{\LV}[1]{\ensuremath{\textrm{\sffamily\bfseries{#1}}}}
\newcommand{\LVq}[1]{\ensuremath{\overline{\textrm{\sffamily\bfseries{#1}}}} }
\newcommand{\LVh}[1]{\hat{\text{\sffamily\bfseries{#1}}}}
\newcommand{\LVt}[1]{\tilde{\text{\sffamily\bfseries{#1}}}}

\newcommand{\tn}{t_{n}}

\newcommand{\realSetDim}[1]{\ensuremath{\mathbb{R}^{#1}}}
\newcommand{\argmin}[1]{\ensuremath{{\text{arg\,min}}_{#1}\,}}
\DeclareSIUnit{\px}{px}
\newtheorem{remark}{Remark}

\title{Projection-based coupling of infrared thermography and stereocorrelation-based digital image correlation}

\date{}

\author{
    \href{https://orcid.org/0000-0002-4999-4558}{\includegraphics[scale=0.06]{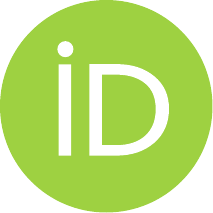}\hspace{1mm}Jendrik-Alexander Tröger} \\
	Institute of Applied Mechanics\\
	Clausthal University of Technology\\
	Adolph-Roemer-Stra{\ss}e 2a \\
	38678 Clausthal-Zellerfeld, Germany \\
	\texttt{jendrik-alexander.troeger@tu-clausthal.de} \\
    \And
    Lutz Müller-Lohse \\
	Institute of Applied Mechanics\\
	Clausthal University of Technology\\
	Adolph-Roemer-Stra{\ss}e 2a \\
	38678 Clausthal-Zellerfeld, Germany \\
	\texttt{lutz.mueller-lohse@tu-clausthal.de} \\
	\And
	\href{https://orcid.org/0000-0003-1849-0784}{\includegraphics[scale=0.06]{orcid.pdf}\hspace{1mm}Stefan Hartmann} \\
	Institute of Applied Mechanics\\
	Clausthal University of Technology\\
	Adolph-Roemer-Stra{\ss}e 2a \\
	38678 Clausthal-Zellerfeld, Germany \\
	\texttt{stefan.hartmann@tu-clausthal.de} \\
}

\renewcommand{\headeright}{}
\renewcommand{\undertitle}{}
\renewcommand{\shorttitle}{Projection-based coupling of infrared thermography and stereocorrelation-based digital image correlation}

\hypersetup{
pdftitle={Projection-based coupling of infrared thermography and stereocorrelation-based digital image correlation},
pdfsubject={},
pdfauthor={Jendrik-Alexander Tröger, Lutz Müller-Lohse, Stefan Hartmann},
pdfkeywords={infrared thermography, digital image correlation, radial basis functions, interpolation, temperature gradient, temperature rate},
hidelinks
}

\begin{document}
\maketitle

\begin{abstract}
Full-field measurement techniques such as digital image correlation and infrared thermography are prevalent in experimental solid mechanics. Digital image correlation is used to analyze surface deformation, while infrared thermography quantifies surface temperature fields. However, sophisticated procedures are necessary to express both datasets in the same Lagrangian frame, especially when analyzing non-flat surfaces. In this study, we propose an external projection-based coupling that uses the pinhole camera model to relate two-dimensional temperature data measured by infrared thermography to three-dimensional point coordinates from stereocorrelation-based digital image correlation. Unlike existing multiview approaches, we utilize two independently calibrated industrial-grade systems and augment the experimental evaluation with the pinhole camera model. The projection matrix of the camera model is calibrated using a single image of a reference object. Through this projection, temperature fields are accurately represented at material points. Our method is particularly suited for, but not restricted to, curved surfaces and is straightforward to embed in existing experimental protocols, as the image registration is kept as is. Additionally, we use radial basis functions as a global interpolation ansatz in both space and time to compute in-plane temperature gradients and even temperature rates on curved surfaces, thereby providing an extensive and information-rich full-field dataset.
\end{abstract}


\section{Introduction}
\label{sec:intro}

Full-field measurement methods have emerged as indispensable tools for generating extensive and information-rich datasets to investigate material behavior under various conditions \citep{hildroux2025}. These datasets serve as a foundation for tasks such as constitutive model calibration \citep{avriletal2008,chenetal2024,roemerhartmanntroegerantonwesselsflascheldelorenzis2025}, model discovery \citep{roemerhartmanntroegerantonwesselsflascheldelorenzis2025,abbasietal2026}, and validation of numerical models \citep{kovscaetal2026,troegersartortigarhuomduesterhartmann2024}. Among the various full-field measurement methods, consider \citep{grediachild2013} for a comprehensive overview, digital image correlation (DIC) using visible light cameras is the most established method for investigating surface deformations \citep{suttonorteuschreier2009,sutton2013,hildroux2006}. Apart from that, infrared thermography (IRT) is an optical, non-destructive method that detects the infrared radiation inevitably emitted by materials, enabling the determination of surface temperature fields \citep{chrysochoos2012,ibarracastanedoetal2013}. Specific applications of full-field temperature data are constitutive model calibration and validation of both isotropic \citep{rosemenzel2020} and anisotropic materials \citep{troegerhartmann2022,troegerkulozikhartmann2025,spenceretal2026}, or even the detection of defects \citep{maierhoferetal2018}. 

However, coupling DIC and IRT presents a challenge due to the inherently different measurement principles. While (stereocorrelation-based) DIC provides (three-dimensional) material point coordinates in a Lagrangian setting, IRT determines pixel temperatures in an Eulerian setting, i.e., two-dimensional spatial information. As a result, it is of particular interest to express both displacement and temperature fields within the same Lagrangian frame (i.e., at a material point in the context of continuum mechanics). For planar specimens, different experimental protocols have been developed in the past. Often, thin specimens are used, where the strain and thermal gradients can be neglected in the thickness direction. This assumption motivates measuring the displacement and temperature fields at opposing sides of the specimen. Related studies include, among others, \citep{chrysochoosetal2010,toussaintetal2012,daietal2015,wangetal2016,jungetal2019,songtrivedisiviour2023,grednevetal2022,jonesetal2024}. In contrast, other studies investigated the same surface using a coupled setup of IRT and 2D-DIC, see, for instance, \citep{bodelotetal2011,nowakmaj2018,yuanwang2019}. In these studies, surface preparation poses a challenge, since DIC requires a pattern for point tracking, whereas IRT performs best on surfaces with homogeneous and high emissivity. Additionally, the experimental setups are typically more involved, as they include dichroic mirrors \citep{bodelotetal2011} or longpass IR filters \citep{nowakmaj2018}. A different approach is to use a single IR camera and perform image correlation on the IR images, a method termed \textit{infrared image correlation} by \citep{maynadieretal2012}. Further related studies using a single IR camera are provided by \citep{silvaravichandran2011,wangliujiang2017} and more recently \citep{chenliuyuyupan2026,qinhuzouyu2025}. Other works employ the near-infrared wavelength band and a single CMOS camera \citep{jailinetal2022}. Instead of coupling DIC and IRT, a notable technique has been developed by \citep{jonesjoneswinters2022} that combines thermographic phosphor with DIC.

The majority of the studies mentioned above focus on combining 2D-DIC with IRT. However, when investigating complex geometries or out-of-plane displacements, 2D-DIC becomes impractical. Stereocorrelation of image pairs captured by two visible light cameras is the most established technique for 3D-DIC. It is noteworthy, though, that a single camera combined with mirrors is also applicable, see \citep{besnardetal2010,yupan2016,panyuzhang2017} for example. The calibration of stereocorrelation-based DIC requires either multiple images of a planar calibration target at different positions and angles \citep{beaubieretal2014} or a 3D target, which necessitates only a single image \citep{besnardetal2010}. Note that the 3D target could be the specimen itself, since its shape is known \citep{beaubieretal2014}. A particular study that couples 3D-DIC and IRT systems is presented in \citep{cholewaetal2016}, wherein the authors employ the pinhole camera model for projection, in conjunction with a model that accounts for image distortion. The approach by \citep{cholewaetal2016} is based on selecting one CCD camera as the primary camera, to which all other images are projected for image registration.

The hybrid multiview framework proposed by \citep{charbaletal2016} can be seen as a more mature approach, as it provides a unified framework for processing image information from visible light and IR cameras. Specific applications in thermo-mechanical fatigue tests are reported by \citep{wangcharbalhildrouxvincent2019,wangcharbaldufourhildrouxvincent2020}. Recently, an extension of this framework toward multimodal correlation analyses, incorporating even tomography data, has been published \citep{zaplaticetal2025}. According to \citep{zaplaticetal2025}, thermo-mechanical coupling or damage leads to digital level variations, making the inclusion of IR images in DIC schemes challenging. As a remedy, brightness and contrast corrections could be applied \citep{charbaletal2016,sciutietal2021}. It should be noted that semi-hybrid multiview approaches, in which IR image information is not included in the DIC scheme, are also possible, see \citep{zaplaticetal2023} for details. The aforementioned studies generally utilize image information directly, specifically the scalar gray levels for DIC and digital levels for IRT. This \textit{eikological} approach, as designated in \citep{hildroux2025}, facilitates convenient \textit{global} formulations of calibration and correlation, even in multimodal and multiview contexts. However, it must be considered that many commercial software and systems are based on subset-based DIC, i.e., \textit{local} DIC in the sense of \citep{hildroux2012}. Thus, coupling stereocorrelation-based DIC and IRT to obtain thermo-mechanical full-field data for model calibration or validation is of particular interest to researchers who do not develop or have access to the core of DIC codes. This interest has motivated earlier work in which three-dimensional point coordinates and pixel temperatures were coupled based on geometrical considerations \citep{hartmannmuellerlohsetroeger2023}. However, this procedure comes at the cost of reduced projection accuracy compared to using a camera model, which is resolved in the present work.

Interpolation techniques are employed to compute strain fields from point coordinates. For that, radial basis functions (RBFs) are utilized by \citep{daietal2015,grothetal2022} for interpolation of 2D-data. A more general formulation for three-dimensional surfaces is presented in \citep{hartmannmuellerlohsetroeger2021}, enabling the computation of principal strain directions through a global interpolation ansatz. Note that other functions, such as B-splines or non-uniform rational B-splines (NURBS), are applicable as well. In this context, \citep{lehmannihlemann2022,lehmannihlemann2025} employ B-splines for interpolation and smoothing of DIC data, whereas \citep{rethoreetal2009} utilize NURBS for the image correlation and subsequent steps directly.

In this work, we present an approach to couple the information from IRT and stereocorrelation-based DIC, namely the pixel temperatures and three-dimensional point coordinates, based on the pinhole camera model. This methodology is similar to earlier studies \citep{cholewaetal2016} or the hybrid multiview framework \citep{charbaletal2016,zaplaticetal2023}. In contrast to those existing studies, we use two independently calibrated industrial-grade systems for IRT and DIC. Consequently, the developed methodology is applicable independently of the system manufacturer or access to DIC codes for calibration and correlation, as we rely entirely on (three-dimensional) point coordinates and pixel temperatures. Therewith, we provide a modular approach to generating extensive and information-rich data for the calibration and validation of thermo-mechanically coupled constitutive models. Our methodology is particularly suited for, but not restricted to, analyzing curved surfaces. The projection matrix of the pinhole camera model is calibrated using a single image of a three-dimensional reference object, which does not have to be heated, in contrast to the procedures by \citep{cholewaetal2016,zengetal2023}. Additionally, we extend the global interpolation approach of \citep{hartmannmuellerlohsetroeger2023} into the temporal domain to evaluate not only temperature gradients on three-dimensional curved surfaces but also the temperature rate, which, to the best of the authors' knowledge, has not yet been addressed in the literature using full-field data.

In Sec.~\ref{sec:projection_interp}, we begin by briefly introducing the projection procedure based on a camera model, which is essential for coupling the two independently calibrated camera systems, here, stereocorrelation-based DIC and IRT. Furthermore, the global interpolation procedure in space and time employing RBFs is explained alongside the computation of in-plane temperature gradients and temperature rates. In Sec.~\ref{sec:experiment}, the proposed methodology is demonstrated and investigated for a thermo-mechanical experiment of a tension-torsion test on a tube-like specimen.

\section{Projection and Interpolation Procedure}
\label{sec:projection_interp}
\begin{figure}[ht]
    \centering
    \includegraphics[width=0.975\textwidth]{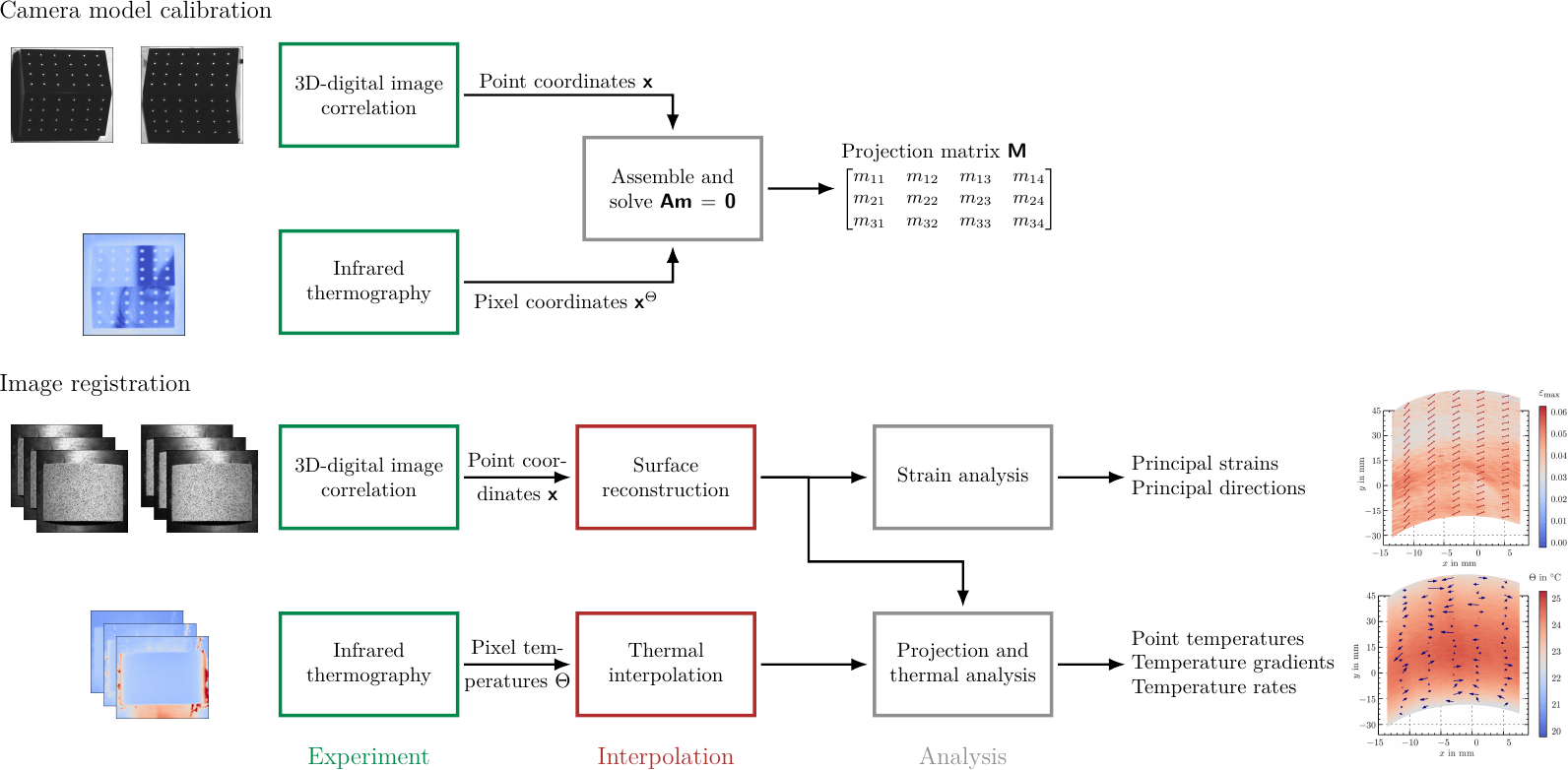}
    \caption{Overview of projection and interpolation procedure to couple independently calibrated stereocorrelation-based digital image correlation and infrared thermography systems. Camera model calibration is performed using a single image of an open book-shaped reference object. Point coordinates $\LV{x}$ are determined by DIC, while pixel coordinates $\LV{x}^\Theta$ are obtained from the IR image via edge detection. Assembling and solving a homogeneous linear system yields the projection matrix $\LM{M}$ of the pinhole camera model. The image registration comprises the actual image acquisition and evaluation. Interpolation is performed for point coordinates $\LV{x}$ and pixel temperatures $\Theta$. The point coordinates $\LV{x}$ are projected into the IR image plane using the projection matrix $\LM{M}$ to evaluate point temperatures on an arbitrarily shaped surface.}
    \label{fig:scheme}
\end{figure}

A camera model describes the projection of a three-dimensional point onto the two-dimensional image plane. This camera model serves as the foundation for coupling measurement data from our stereocorrelation-based DIC system and IR camera, which is briefly explained in Sec.~\ref{sec:projection}. Subsequently, the individual relations for in-plane temperature gradient and temperature rate in a general curvilinear setting are derived in Sec.~\ref{sec:tempGrad_tempVeloc}. Finally, the global interpolation ansatz employing RBFs is presented in Sec.~\ref{sec:interpRBF}, along with the specific relations for computing the in-plane temperature gradient and the temperature rate. The procedure during both camera model calibration and image registration and evaluation is depicted in Fig.~\ref{fig:scheme}.

\subsection{Projection using a Camera Model}
\label{sec:projection}
Camera models enable accurate mapping of three-dimensional object points to two-dimensional image planes. This mapping is the basis for optical imaging techniques, such as DIC. In contrast to DIC, where the individual projections of object points in the captured images are of interest, we employ the projection to obtain temperatures and point coordinates in the same Lagrangian frame. We use the pinhole camera model as a linear transformation without accounting for image distortion. Neglecting image distortion is justified in our experimental setup and is discussed in Sec.~\ref{sec:exp_projAcc} in more detail. The following explanations are restricted to the scope required for this work. Further details about camera models are provided by \citep{faugeras1993,suttonorteuschreier2009,hartleyzisserman2004}, among others. A comprehensive review of various calibration techniques for both linear and nonlinear models is available in \citep{salviarmanguebatlle2002}.

In our case, we obtain pixel temperatures $\Theta_i(t_n)$, $i = 1,\ldots,n_\mathrm{IR}$, from IRT and three-dimensional point coordinates $\vec{x}_j(t_n)$, $j = 1,\ldots,n_\mathrm{DIC}$, from stereocorrelation-based DIC. Of course, the image capturing of both systems is synchronized, i.e., the time points $t_n$, $n = 0,\ldots,n_\mathrm{t}$, coincide up to a small trigger delay, which is accounted for during temporal interpolation. In the following, we will always refer to a Cartesian basis and formulate the point coordinates as column vectors $\LV{x}\in\mathbb{R}^3$ instead of geometrical vectors $\vec{x}\in\mathbb{V}^3$. The image plane is given by the IRT measurement and is based on the pixel coordinates $\LV{x}^\Theta = \lbrace u^\Theta, v^\Theta \rbrace^\top$, which read in homogeneous coordinates $\LVq{x}^\Theta = \lbrace \sigma u^\Theta, \sigma v^\Theta, \sigma \rbrace^\top$ with $\sigma$ being a scale factor. The homogeneous coordinates of any three-dimensional object point are $\LVq{x} = \lbrace \LV{x}, 1 \rbrace^\top = \lbrace x, y, z, 1 \rbrace^\top$ and are determined with DIC. Our approach is based on the perspective projection equation
\begin{equation}
    \label{eq:perspectiveProjEq}
    \LVq{x}^\Theta = \LM{M}\,\LVq{x},
\end{equation}
where $\LM{M}\in\mathbb{R}^{3 \times 4}$ is the projection or transformation matrix, incorporating the elementary operations between world and camera coordinate systems, see \citep{suttonorteuschreier2009,hartleyzisserman2004} for details. Before image registration, the coefficients of the projection matrix $\LM{M}$ must be calibrated. These coefficients include both intrinsic (i.e., internal camera parameters) and extrinsic (i.e., the camera's angle and position) parameters. 

\begin{figure}[ht]
    \centering
    \begin{subfigure}[b]{0.3\textwidth}
        \centering
        \includegraphics[height=0.8\textwidth]{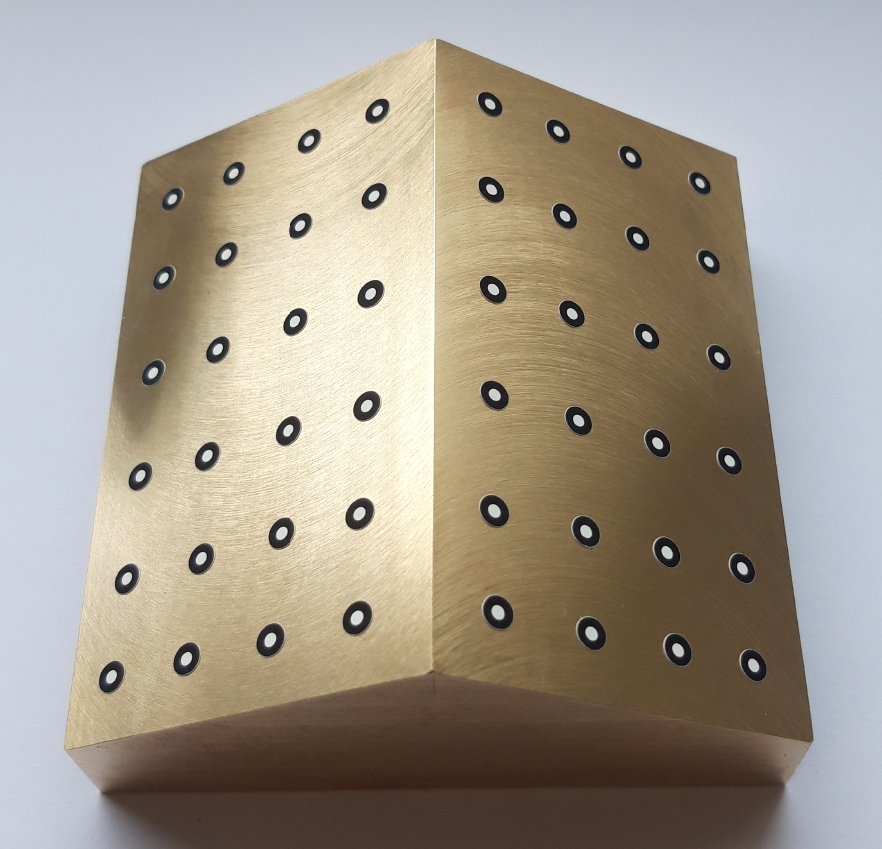}
        \caption{}
        \label{fig:refObj_image}
    \end{subfigure}
    \hspace{0.025\textwidth}
    \begin{subfigure}[b]{0.3\textwidth}
        \centering
        \includegraphics[height=0.8\textwidth]{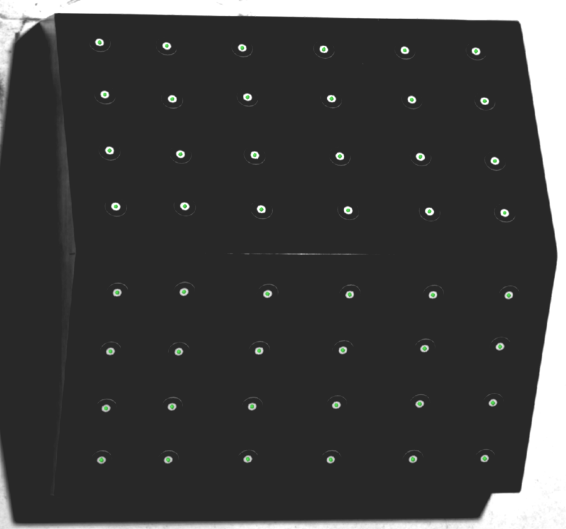}
        \caption{}
        \label{fig:refObj_DIC}
    \end{subfigure}
    \hspace{0.025\textwidth}
    \begin{subfigure}[b]{0.3\textwidth}
        \centering
        \includegraphics[height=0.8\textwidth]{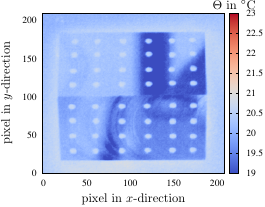}
        \caption{}
        \label{fig:refObj_IR}
    \end{subfigure}
    \caption{Reference object for camera model calibration. (a) open book-shaped reference target with point markers, (b) center points of markers indicated by green dots after image correlation (shown in left camera image), (c) IR image of reference object.}
    \label{fig:refObj}
\end{figure}

\paragraph{Calibration} Since the coefficients of the transformation are unknown, we employ the Direct Linear Transformation (DLT) using a single image of a three-dimensional reference object. Such reference objects for DIC are typically open-book targets with a chessboard pattern \citep{besnardetal2010}. In this work, we draw on an open book-shaped reference object with several points visible in the two visible light cameras and the IR camera, see Fig.~\ref{fig:refObj}.
The point markers are captured by DIC, returning the point coordinates $\LV{x}$ of their centers (Fig.~\ref{fig:refObj_DIC}). Since the reference object is made of brass, the point markers have a significantly different emissivity and are therefore also observable in the temperature data of the IR camera, see Fig.~\ref{fig:refObj_IR}. The pixel coordinates $\LV{x}^\Theta$ of each marker's center are determined using an intensity-based approach computing the center of mass. In general, different methods are available to determine the center of the markers in the IR image. To determine the most suitable method for the present setting of a comparatively low resolution in the IR image, a benchmark analysis of different center localization methods is provided in \ref{app:marker_detection_IR}. In the literature, heated reference targets are employed, see for example \citep{charbaletal2016} and \citep{zengetal2023}. As shown in \ref{app:marker_detection_IR}, even an unheated reference object yields subpixel accuracy when using the intensity-based center of mass computation. Thus, we omit any prior heating of the reference target for the camera model calibration in this study.

By denoting the rows of the projection matrix $\LM{M}$ as $\LV{m}_k^\top$, $\LV{m}_k\in\mathbb{R}^4$, $k=1,2,3$, the scale factor reads $\sigma = \LV{m}_3^\top \LVq{x}_i$, $i=1,\ldots,n_\mathrm{cal}$. The image of the three-dimensional reference objects provides $n_\mathrm{cal}$ pairs of coordinates, $\LVq{x}_i^\Theta$ from IRT in the image plane and $\LVq{x}_i$ from DIC as object points, $i = 1,\ldots,n_\mathrm{cal}$. Eq.~\eqref{eq:perspectiveProjEq} provides two equations per reference point,
\begin{equation}
    \label{eq:DLT}
    \renewcommand{\arraystretch}{1.25}
    \begin{Bmatrix}
    u_i^\Theta \\ v_i^\Theta
    \end{Bmatrix}
    = 
    \begin{Bmatrix}
        {\LV{m}_1^\top \LVq{x}_i}/{\LV{m}_3^\top \LVq{x}_i} \\
        {\LV{m}_2^\top \LVq{x}_i}/{\LV{m}_3^\top \LVq{x}_i}        
    \end{Bmatrix}
    \quad \rightarrow \quad
    \renewcommand{\arraystretch}{1.25}
    \begin{bmatrix}
        \LVq{x}_i^\top & \LV{0}^\top & -u_i^\Theta \LVq{x}_i^T \\
        \LV{0}^\top & \LVq{x}_i^\top & -v_i^\Theta \LVq{x}_i^T \\
    \end{bmatrix}
    \underbrace{
    \begin{Bmatrix}
        \LV{m}_1 \\ \LV{m}_2 \\ \LV{m}_3 
    \end{Bmatrix}
    }_{\LV{m}}    
    =
    \LV{0}.
\end{equation}
While Eq.~\eqref{eq:DLT}$_1$ is nonlinear in the sought coefficients of the projection matrix, the reformulation Eq.~\eqref{eq:DLT}$_2$ provides a linear relationship, which is beneficial for the calibration \citep{hartleyzisserman2004}. Since every reference point provides two equations, at least six reference points are required to determine the coefficients of the projection matrix $\LM{M}$, which are arranged as column vector $\LV{m} = \lbrace m_{11}, m_{12}, m_{13}, m_{14}, m_{21}, m_{22}, m_{23}, m_{24}, m_{31}, m_{32}, m_{33}, m_{34} \rbrace^\top$, $\LV{m}\in\mathbb{R}^{12}$, in Eq.~\eqref{eq:DLT}$_2$. The equations arising from $n_\mathrm{cal}$ reference points can be row-wise compiled in a matrix,
\begin{equation}
    \label{eq:coeffMat_homogenLinSys}
    \renewcommand{\arraystretch}{1.25}
    \LM{A} = \begin{bmatrix}
        \LVq{x}_1^\top & \LV{0}^\top & -u_1^\Theta \LVq{x}_1^\top \\
        \LV{0}^\top & \LVq{x}_1^\top & -v_1^\Theta \LVq{x}_1^\top \\
        \vdots & \vdots & \vdots \\
        \LVq{x}_{n_\mathrm{cal}}^\top & \LV{0}^\top & -u_{n_\mathrm{cal}}^\Theta \LVq{x}_{n_\mathrm{cal}}^\top \\
        \LV{0}^\top & \LVq{x}_{n_\mathrm{cal}}^\top & -v_{n_\mathrm{cal}}^\Theta \LVq{x}_{n_\mathrm{cal}}^\top
    \end{bmatrix},
\end{equation}
$\LM{A}\in\mathbb{R}^{2n_\mathrm{cal} \times 12}$, that ultimately leads to a homogeneous system of linear equations
\begin{equation}
    \label{eq:homogenLinSys}
    \LM{A} \LV{m} = \LV{0}. 
\end{equation}
In addition to the preceding explanations, we apply the normalization of both image coordinates $\LVq{x}^\Theta$ and point coordinates $\LVq{x}$ as explained by \citep[pp. 180-181] {hartleyzisserman2004} before constructing $\LM{A}$ and solving the homogeneous system~\eqref{eq:homogenLinSys}. Since we use more than six reference points for accurate calibration of the projection between DIC and IRT data, the DLT minimizes the error in a least-squares sense,
\begin{equation}
    \label{eq:leastSquares}
    \LV{m}^* = \argmin{\LV{m}} \phi(\LV{m})
    \quad \text{s.t.} \quad
    \vert\vert\LV{m}\vert\vert = 1
    \quad \text{with} \quad
    \phi(\LV{m}) = \frac12 \vert\vert \LV{r} \vert\vert^2 = \frac12 \vert\vert \LM{A}\LV{m} \vert\vert^2.
\end{equation}
Because the system~\eqref{eq:homogenLinSys} is linear, linear least-squares or singular value decomposition can be used to compute the coefficients of the projection matrix $\LM{M}$, where we choose the latter, see the explanations in \citep[App. B]{suttonorteuschreier2009} and \citep[App. 5]{hartleyzisserman2004} as well. The linear solution can be used as an initial estimate to further reduce geometric error in the image by iterative techniques. However, \citep[p. 182]{hartleyzisserman2004} demonstrate that this step often yields insignificant improvements and it is therefore omitted in this work. It is worth mentioning that \citep{baronemarrazzooton2020} introduced a weighted variant of the DLT, claiming more robust and precise solutions in the presence of anisotropic uncertainties.

\begin{remark}
It should be emphasized that the stereocorrelation-based DIC system is independently calibrated using multiple images of a planar reference object following established practice. The IR camera is calibrated by the manufacturer for the temperature range $\num{-40}\ldots\SI{120}{\celsius}$. As a result, the reference object in Fig.~\ref{fig:refObj} serves only for calibrating the pinhole camera model, specifically the coefficients of the projection matrix $\LM{M}$. As a result, our approach differs from the hybrid multiview framework introduced by \citep{charbaletal2016} and its semi-hybrid variant \citep{zaplaticetal2023}, since we employ individually calibrated imaging systems for IRT and stereocorrelation-based DIC. This external calibration of the projection, in a certain sense, might offer less accuracy than the hybrid multiview framework because we operate with noisy data -- point coordinates $\LV{x}$ and pixel temperatures $\Theta$ -- rather than using the image information. However, our approach allows straightforward coupling of data from industrial-grade and independently calibrated imaging systems, which is of particular interest in solid mechanics for purposes such as constitutive model calibration or validation of numerical simulations.
\end{remark}

\paragraph{Image Acquisition} Essentially, Eq.~\eqref{eq:perspectiveProjEq} enables the assignment of temperature values (measured in an Eulerian setting using IRT) to material points whose positions are obtained from DIC in a Lagrangian setting. If solely the temperatures at material points on curved surfaces are of interest, calibrating the projection matrix $\LM{M}$ using a reference object, followed by interpolating the pixel temperatures and applying the projection equation \eqref{eq:perspectiveProjEq}, is sufficient. However, our objective is to compute spatial temperature gradients and temperature rates, requiring a continuous surface description from discrete pointwise data, as explained in the following subsection.

\subsection{Computation of Temperature Gradient and Temperature Rate in Curved Surfaces}
\label{sec:tempGrad_tempVeloc}
In general, the reconstruction of a curvilinear surface from discrete pointwise data is achieved using surface parameters $\Psi^1$ and $\Psi^2$, $\bm{\Psi} = \lbrace \Psi^1, \Psi^2 \rbrace$, see \ref{app:curvedKinematics} for details. Coupling stereocorrelation-based DIC and IRT enables the assignment of temperatures to material points on a curved surface. However, this assignment alone does not yet reveal all the information provided by such full-field measurements, which typically capture surface deformation and temperature over a certain time interval during loading or unloading. Of particular interest are the spatial temperature gradient within the curved surface and the temperature rate, both of which remain unavailable when using IRT alone. Consequently, in our coupled setup, the temperature $\Theta = \hat{\Theta}\left(\LVh{x}^\Theta\left(\LVh{x}(\bm{\Psi},t)\right),t\right)$ depends implicitly on the real-world coordinates $\LV{x}$ of material points on the curved surface, which themselves depend on the surface parameters $\bm{\Psi}$ and time $t$. Note that we use a hat $\hat{\langle\bullet\rangle}$ to denote functions, thereby distinguishing between functions and arguments. Additionally, Greek letters are counted from 1 to 2.

\paragraph{Temperature Gradient} According to the derivations in \citep{hartmannmuellerlohsetroeger2021}, the in-plane temperature gradient on an arbitrarily curved surface can be expressed as
\begin{equation}
    \label{eq:tempGrad}
    \widehat{\grad}\,\Theta = \dif{\Theta}{\Psi^\alpha} \LV{a}^\alpha,
\end{equation}
where $\LV{a}^\alpha$ denotes the gradient vector in the current configuration. Applying the chain rule, the derivative of the temperature with respect to the surface parameters reads
\begin{equation}
    \label{eq:dThetadPsi}
    \dif{\Theta}{\Psi^\alpha} = \dif{\hat{\Theta}}{\LV{x}^\Theta} \difn{\LVh{x}^\Theta}{\LV{x}} \dif{\LVh{x}}{\Psi^\alpha}.
\end{equation}
The partial derivative $\partial\hat{\Theta}/\partial\LV{x}^\Theta$ is obtained from a continuous interpolation of the pixel-wise temperature data with respect to the pixel coordinates. Similarly, the last term in Eq.~\eqref{eq:dThetadPsi}, $\partial\LVh{x}/\partial\Psi^\alpha$, follows from the continuous description of the curved surface using surface parameters $\bm{\Psi}$, see \ref{app:curvedKinematics} as a brief introduction. On the contrary, the total derivative $\D\LVh{x}^\Theta/\D\LV{x}$, $\D\LVh{x}^\Theta/\D\LV{x}\in\realSetDim{2\times 3}$,  is computed by differentiating the projection equation \eqref{eq:perspectiveProjEq}. This derivative is given by
\begin{equation}
    \label{eq:derivProjection}
    \renewcommand{\arraystretch}{1.25}
    \difn{\LVh{x}^\Theta}{\LV{x}} = \frac{1}{\LVt{m}_3^\top\LV{x}+m_{34}}\left(
    \begin{bmatrix}
    \LVt{m}_1^\top \\ \LVt{m}_2^\top
    \end{bmatrix}
    - \LV{x}^\Theta \LVt{m}_3^\top
    \right),
\end{equation}
where $\LVt{m}_k\in\mathbb{R}^3$, $k=1,2,3$, denote the partial row vectors $\LVt{m}_k^\top = \lbrace m_{k1},m_{k2},m_{k3}\rbrace^\top$ of the projection matrix $\LM{M}$. For brevity, the detailed derivation of the derivative~\eqref{eq:derivProjection} is provided in \ref{app:derivativeProjection}.

\paragraph{Temperature Rate} In this work, we are interested in evaluating the temperature rate within the curved surface as well. The temperature rate is defined as the material time derivative of the temperature,
\begin{equation}
    \label{eq:tempVeloc}
    \dot{\Theta} = \difn{\hat{\Theta}\left(\LVh{x}^\Theta\left(\LVh{x}(\bm{\Psi},t)\right),t\right)}{t} = \dif{\hat{\Theta}}{t} + \dif{\hat{\Theta}}{\LV{x}^\Theta} \difn{\LVh{x}^\Theta}{\LV{x}} \dif{\LVh{x}}{t}.
\end{equation}
The temperature rate consists of a local part, $\partial\hat{\Theta}/\partial t$, and a convective contribution. Once again, a continuous description, i.e., interpolation, of the pixel temperatures is necessary, now with respect to both time $t$ and pixel coordinates $\LV{x}^\Theta$. Compared to the temperature gradient~\eqref{eq:dThetadPsi}, only the last derivative in the second part of Eq.~\eqref{eq:tempVeloc} differs, as it now involves the velocity of the coordinates $\LV{x}$. Hence, the relation~\eqref{eq:derivProjection} for the derivative of the projection is employed again.

Up to this point, all relations are formulated in a general setting. The in-plane temperature gradient $\widehat{\grad}\,\Theta$ and the temperature rate $\dot{\Theta}$ can be determined from coupled DIC and IRT data, provided that continuously differentiable interpolations of the pixel temperatures $\Theta$ (with respect to time $t$ and pixel coordinates $\LV{x}^\Theta$) and of the point coordinates $\LV{x}$ (with respect to time $t$ and surface parameters $\bm{\Psi}$) are available. Before this, the projection matrix $\LM{M}$ must be calibrated using a reference object to enable the assignment of temperature values to material points. In what follows, we specify these relations for RBF-based interpolation.

\subsection{Interpolation with Radial Basis Functions}
\label{sec:interpRBF}
In this work, we utilize RBFs to reconstruct surfaces via interpolation. The $k$-th RBF $\hat{m}(\rho_k)$ is defined at so-called center points $\bm{\Psi}_k$ using a normalized distance function $\rho_k = \hat{\rho}(\bm{\Psi},\bm{\Psi}_k) = {R}/{R_0} = {\vert\vert \bm{\Psi} - \bm{\Psi}_k \vert\vert}/{R_0}$ with $R_0 = \SI{1}{\mm}$. While different RBFs could be used to represent distinct regions of the curvilinear surface, in the present work, we employ the same RBF $\hat{m}$ across the entire spatial and temporal domain. The specific properties of RBFs are discussed in considerable detail by \citep{buhmann2004,biancolini2017}. Using RBFs, a point on a curvilinear surface is given by
\begin{equation}
    \label{eq:x_RBFInterp}
    \LV{x}(\bm{\Psi},t) = \sum_{j=1}^3 \left(\sum_{k=1}^{n_\mathrm{RBF}^\mathrm{s}} \hat{m}(\hat{\rho}(\bm{\Psi},\bm{\Psi}_k))A_{kj}^\mathrm{s}(t) + \sum_{l=1}^{n_\mathrm{Mon}^\mathrm{s}} \hat{n}_l(\bm{\Psi})B_{lj}^\mathrm{s}(t)\right)\LV{e}_j.
\end{equation}
The coefficients $A_{kj}^\mathrm{s}$ and $B_{lj}^\mathrm{s}$ are obtained by solving a system of linear equations in the case of interpolation or by solving a linear least-squares problem when performing regression, \citep{hartmannmuellerlohsetroeger2021}. During interpolation, the center points $\bm{\Psi}_k$ are chosen to match the curvilinear convective coordinates of the data points. Here, the curvilinear convective coordinates are obtained from a stationary stereocorrelation-based DIC measurement. From this measurement, the mean value of each point coordinate is determined, and subsequently, these points are projected onto a plane, $\LV{x}_k \mapsto \bm{\Psi}_k$. In Eq.~\eqref{eq:x_RBFInterp}, we include also monomials
\begin{equation}
    \label{eq:monomials_space}
    \LVh{n}^\top(\bm{\Psi}) = \lbrace \hat{n}_1(\bm{\Psi}),\hat{n}_2(\bm{\Psi}), \hat{n}_3(\bm{\Psi})\rbrace^\top = \lbrace 1, \: \Psi^1, \: \Psi^2\rbrace^\top.
\end{equation}
In contrast to earlier works, \citep{hartmannmuellerlohsetroeger2021,hartmannmuellerlohsetroeger2023}, the interpolation is extended to the temporal domain, enabling the computation of temperature rates. Consequently, the coefficients in Eq.~\eqref{eq:x_RBFInterp} become time-dependent and are interpolated with RBFs as well,
\begin{align}
    \label{eq:timeDependent_RBFcoeffs_A}
    A_{kl}^\mathrm{s}(t) &= \sum_{i=1}^{n_\mathrm{RBF}^\mathrm{t}} \hat{m}(\tilde{\rho}(t,\tilde{t}_i)) a_i^\mathrm{t} + \sum_{j=1}^{n_\mathrm{Mon}^\mathrm{t}} \tilde{n}_j(t) \tilde{a}_j^\mathrm{t} \\
    \label{eq:timeDependent_RBFcoeffs_B}
    B_{lm}^\mathrm{s}(t) &= \sum_{i=1}^{n_\mathrm{RBF}^\mathrm{t}} \hat{m}(\tilde{\rho}(t,\tilde{t}_i)) b_i^\mathrm{t} + \sum_{j=1}^{n_\mathrm{Mon}^\mathrm{t}} \tilde{n}_j(t) \tilde{b}_j^\mathrm{t},
\end{align}
where 
\begin{equation}
    \label{eq:monomials_time}
    \LVt{n}^\top(t) = \lbrace \tilde{n}_1(t),\tilde{n}_2(t) \rbrace^\top = \lbrace 1, \: t \rbrace^\top
\end{equation}
represent monomials again. In the temporal domain, the center points $\tilde{t}_i$ correspond to the time points $t_n$, $n = 1,\ldots,n_\mathrm{t}$, of image capturing. The normalized distance is given as $\tilde{\rho}(t,\tilde{t}_i) = \vert t - \tilde{t}_i \vert/\tilde{R}_0$ with $\tilde{R}_0 = \SI{1}{\s}$. As a result, the coefficients $\LV{a}^\mathrm{t}\in\mathbb{R}^{n_\mathrm{RBF}^\mathrm{t}}$, $\LVt{a}^\mathrm{t}\in\mathbb{R}^{n_\mathrm{Mon}^\mathrm{t}}$, $\LV{b}^\mathrm{t}\in\mathbb{R}^{n_\mathrm{RBF}^\mathrm{t}}$, $\LVt{b}^\mathrm{t}\in\mathbb{R}^{n_\mathrm{Mon}^\mathrm{t}}$ in Eqs.~\eqref{eq:timeDependent_RBFcoeffs_A} and~\eqref{eq:timeDependent_RBFcoeffs_B} must be determined by solving a system of linear equations, see \citep{hartmannmuellerlohsetroeger2021} in the context of spatial interpolation. This procedure enables reconstruction of a continuous surface description from discrete DIC data in both the spatial and temporal domains.

The procedure for the IR data, namely the pixel temperatures, is similar to the previous explanations. In this case, the pixel temperatures in the image plane are interpolated using RBFs,
\begin{equation}
    \label{eq:Theta_RBFInterp}
    \Theta(\LV{x}^\Theta,t) = \sum_{k=1}^{n_\mathrm{RBF}^\Theta} \hat{m}(\check{\rho}(\LV{x}^\Theta,\LV{x}^\Theta_k)) a_k^\Theta(t) + \sum_{l=1}^{n_\mathrm{Mon}^\Theta} \check{n}_{l}(\LV{x}^\Theta) b_l^\Theta(t),
\end{equation}
with
\begin{align}
    \label{eq:monomials_Theta}
    \check{\LV{n}}^\top &= \lbrace \check{n}_1(\LV{x}^\Theta),\check{n}_2(\LV{x}^\Theta),\check{n}_3(\LV{x}^\Theta) \rbrace^\top = \lbrace 1, \: u^\Theta,\: v^\Theta \rbrace^\top, \\
    \label{eq:timeDependent_RBFcoeffs_aTheta}
    a_{k}^\mathrm{\Theta}(t) &= \sum_{i=1}^{n_\mathrm{RBF}^\mathrm{t}} \hat{m}(\tilde{\rho}(t,\tilde{t}_i)) \hat{a}_i^\mathrm{t} + \sum_{j=1}^{n_\mathrm{Mon}^\mathrm{t}} \tilde{n}_j(t) \check{a}_j^\mathrm{t}, \\
    \label{eq:timeDependent_RBFcoeffs_bTheta}
    b_{l}^\mathrm{\Theta}(t) &= \sum_{i=1}^{n_\mathrm{RBF}^\mathrm{t}} \hat{m}(\tilde{\rho}(t,\tilde{t}_i)) \hat{b}_i^\mathrm{t} + \sum_{j=1}^{n_\mathrm{Mon}^\mathrm{t}} \tilde{n}_j(t) \check{b}_j^\mathrm{t}.
\end{align}
Hence, the unknown coefficients $\LVh{a}^\mathrm{t}\in\mathbb{R}^{n_\mathrm{RBF}^\mathrm{t}}$, $\check{\LV{a}}^\mathrm{t}\in\mathbb{R}^{n_\mathrm{Mon}^\mathrm{t}}$, $\LVh{b}^\mathrm{t}\in\mathbb{R}^{n_\mathrm{RBF}^\mathrm{t}}$, and $\check{\LV{b}}^\mathrm{t}\in\mathbb{R}^{n_\mathrm{Mon}^\mathrm{t}}$ in Eqs.~\eqref{eq:timeDependent_RBFcoeffs_aTheta} and~\eqref{eq:timeDependent_RBFcoeffs_bTheta} have to be determined based on the IR data. Note that this interpolation is independent of the previous surface reconstruction, since it is formulated with respect to the pixel coordinates $\LV{x}^\Theta = \lbrace u^\Theta,\: v^\Theta \rbrace^\top$ in the image plane of the IR data. The normalized distance is defined as $\check{\rho}(\LV{x}^\Theta,\LV{x}^\Theta_k) = \vert\vert \LV{x}^\Theta - \LV{x}^\Theta_k \vert\vert/\check{R}_0$ with $\check{R}_0 = \SI{1}{\px}$. Although the image acquisition in both camera systems is synchronized, a small trigger delay inevitably occurs, i.e., there could be a small offset between the time points $\tn$ in DIC and IRT. To avoid a biased temperature rate~\eqref{eq:tempVeloc} due to this trigger delay, the temporal interpolation of the DIC and IRT data is performed at the exact timestamps, which are associated with each image.

Interpolating pixel temperatures is necessary to assign temperature values to material points on the curvilinear surface. After calibrating the projection matrix and performing image acquisition as described in Sec.~\ref{sec:projection}, the point coordinates from DIC are projected into the image plane using the projection equation \eqref{eq:perspectiveProjEq}. However, the resulting pixel coordinates do not necessarily coincide with the integer pixel values $\LV{x}^\Theta$ in the IR data. Therefore, the interpolation \eqref{eq:Theta_RBFInterp} must be evaluated to assign temperature values to material points on the curvilinear surface.

The computation of temperature gradient~\eqref{eq:dThetadPsi} and temperature rate~\eqref{eq:tempVeloc} requires the four derivatives that rely on the RBF-based continuous interpolation in the spatial and temporal domains, 
\begin{align}
    \label{eq:dThetadxTh}
    \dif{\hat{\Theta}}{\LV{x}^\Theta} &= \sum_{k=1}^{n_\mathrm{RBF}^\Theta} \difn{\hat{m}}{\rho} \difn{\check{\rho}}{\LV{x}^\Theta} a_k^\Theta(t) + \sum_{l=1}^{n_\mathrm{Mon}^\Theta} \difn{\check{n}_{l}(\LV{x}^\Theta)}{\LV{x}^\Theta} b_l^\Theta(t), \\
    \label{eq:dThetadt}
    \dif{\hat{\Theta}}{t} &= \sum_{k=1}^{n_\mathrm{RBF}^\Theta} \hat{m}(\check{\rho}) \difn{a_k^\Theta(t)}{t} + \sum_{l=1}^{n_\mathrm{Mon}^\Theta} \check{n}_{l}(\LV{x}^\Theta) \difn{b_l^\Theta(t)}{t}, \\
    \label{eq:dxdPsi}
    \dif{\LVh{x}}{\Psi^\alpha} &= \sum_{j=1}^3 \left(\sum_{k=1}^{n_\mathrm{RBF}^\mathrm{s}} \difn{\hat{m}}{\rho} \dif{\hat{\rho}}{\Psi^\alpha} A_{kj}^\mathrm{s}(t) + \sum_{l=1}^{n_\mathrm{Mon}^\mathrm{s}} \dif{\hat{n}_l}{\Psi^\alpha} B_{lj}^\mathrm{s}(t)\right)\LV{e}_j, \\
    \label{eq:dxdt}
    \dif{\LVh{x}}{t} &= \sum_{j=1}^3 \left(\sum_{k=1}^{n_\mathrm{RBF}^\mathrm{s}} \hat{m}(\hat{\rho}) \difn{A_{kj}^\mathrm{s}(t)}{t} + \sum_{l=1}^{n_\mathrm{Mon}^\mathrm{s}} \hat{n}_l(\bm{\Psi}) \difn{B_{lj}^\mathrm{s}(t)}{t}\right)\LV{e}_j.
\end{align}
A more detailed formulation of the individual derivatives is provided in \ref{app:derivativesRBFInterp} for brevity. The derivative $\D\hat{m}/\D\rho$ depends on the chosen RBF for the interpolation and is typically available as an analytical derivative.

It is important to note that the overall methodology presented in this study is not restricted to RBFs or to the use of the same interpolation functions in both the spatial and temporal domains. Any interpolation function may be employed, provided it yields a continuous surface description and interpolates the temperature field. Here, RBFs are chosen to achieve $C^{\infty}$-continuity during interpolation.

\section{Experiment}
\label{sec:experiment}
The developed projection approach for coupling point coordinates from stereocorrelation-based DIC with pixel temperature data from IRT is demonstrated in a thermo-mechanically coupled setting. Specifically, a tension-torsion test of a tube-like specimen is investigated in detail, inherently necessitating the analysis of full-field data on a curved surface. Evaluating in-plane temperature gradients and temperature rates yields an extensive set of full-field data. Technical details of the utilized camera systems are summarized in \ref{app:imagingSystems}.

\subsection{Experimental Setup}
\label{sec:exp_setup}
\begin{figure}[ht]
    \centering
    \begin{subfigure}[c]{0.52\textwidth}
        \centering
        \includegraphics[width=\linewidth]{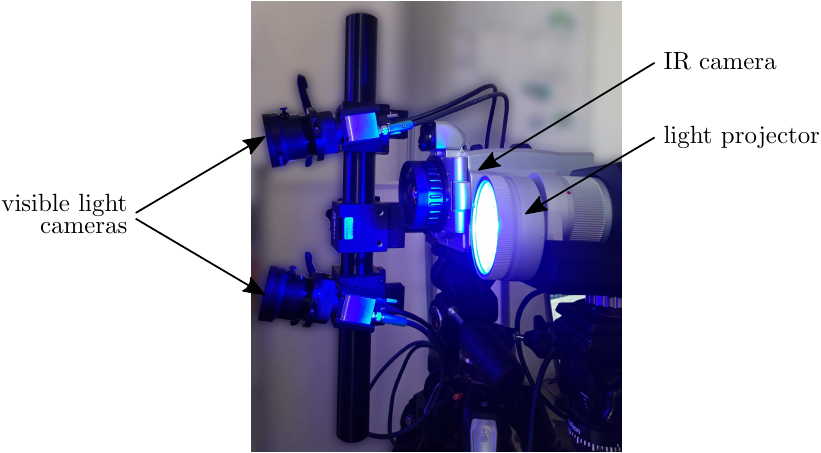}
        \caption{}
        \label{fig:expSetup_thermoMech}
    \end{subfigure}
    \hspace{0.02\textwidth}
    \begin{subfigure}[c]{0.45\textwidth}
        \centering
        \includegraphics[width=\linewidth]{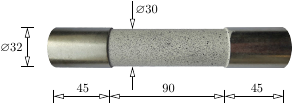}
        \caption{}
        \label{fig:zamak-specimen}
    \end{subfigure}
    \caption{Experimental setup of the tension-torsion test. a) Alignment of the full-field imaging systems. b) Tube-like specimen (dimensions in \si{\mm})}
    \label{fig:expSetup_specimen_zamak}
\end{figure}

The alignment of the imaging systems toward the universal testing machine is visualized in Fig.~\ref{fig:expSetup_thermoMech}.
The measuring distance between the visible light cameras and the specimen surface was approximately $\SI{620}{\mm}$. Measurements are performed on the same specimen surface to enable projection. The surface is therefore varnished with a stochastic speckle pattern. Since IRT, unlike DIC, performs best on homogeneous surfaces with high emissivity, using a speckle pattern is not ideal. However, in \citep{hartmannmuellerlohsetroeger2021} we have shown that for the temperature range under investigation (close to room temperature), the influence of the speckle pattern on the temperature acquisition is negligible if the speckle pattern is sufficiently fine. This assumption holds because the varnish has a comparatively high emissivity ($\varepsilon=\num{0.96}$, determined experimentally under known temperature conditions) and the spatial resolution of the IR camera is significantly coarser than that of the visible light cameras. As a result, the speckle pattern is not visible in the IR images \citep{zaplaticetal2023}. Similar studies employed even higher emissivity values, see \citep{goidescuetal2013,neubaueretal2022}, for example.

\begin{figure}[ht]
    \centering
    \begin{subfigure}[b]{0.3\linewidth}
        \centering
        \includegraphics[width=0.9\linewidth]{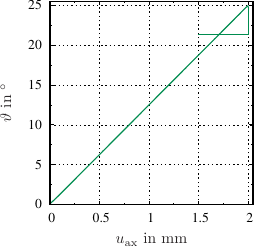}
        \caption{}
        \label{fig:zamak_angle-disp}        
    \end{subfigure}
    \hspace{0.02\linewidth}
    \begin{subfigure}[b]{0.3\linewidth}
        \centering
        \includegraphics[width=0.9\linewidth]{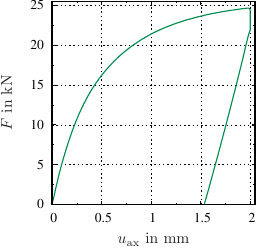}
        \caption{}
        \label{fig:zamak_force}        
    \end{subfigure}
    \hspace{0.02\linewidth}
    \begin{subfigure}[b]{0.3\linewidth}
        \centering
        \includegraphics[width=0.95\linewidth]{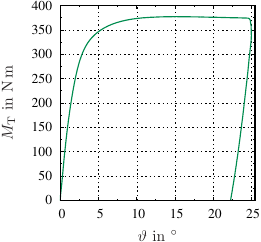}
        \caption{}
        \label{fig:zamak_torque}        
    \end{subfigure}
    \caption{Tension-torsion test on zinc die-casting alloy. a) Prescribed displacement $u_\mathrm{ax}$ and twist angle $\vartheta$. b) Axial force response. c) Torque response.}
    \label{fig:zamak_singleValuedResults}
\end{figure}

The thin-walled tube-like specimen, visualized in Fig.~\ref{fig:zamak-specimen}, is made of a zinc die-casting alloy. The outer dimensions are shown in Fig.~\ref{fig:zamak-specimen}, whereas the inner diameter is \SI{26}{\mm}. The specimen is subjected to a proportional tension-torsion load, followed by an unloading step to zero torque and another to zero axial force, see Fig.~\ref{fig:zamak_angle-disp}.
To introduce a temperature increase due to plastic deformation, the maximum twist angle $\vartheta = \ang{25}$ and axial displacement $u_\mathrm{ax} = \SI{2}{\mm}$ are applied within \SI{5}{\s}, i.e., $\dot{\vartheta} = \SI{5}{\degree\per\s}$ and $\dot{u}_\mathrm{max} = \SI{0.4}{\mm\per\s}$. Images in both DIC and IRT were synchronously acquired at $\SI{2}{\Hz}$.

\subsection{Projection Accuracy}
\label{sec:exp_projAcc}
\begin{figure}[htb]
    \centering
    \begin{subfigure}[b]{0.45\linewidth}
        \centering
        \includegraphics[width=0.75\linewidth]{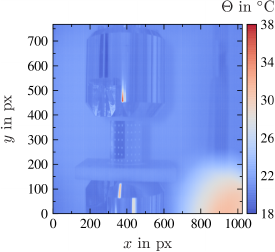}
        \caption{}
        \label{fig:refObj_fullIRview}
    \end{subfigure}
    \hspace{0.04\textwidth}
    \begin{subfigure}[b]{0.45\linewidth}
        \centering
        \includegraphics[width=0.8\linewidth]{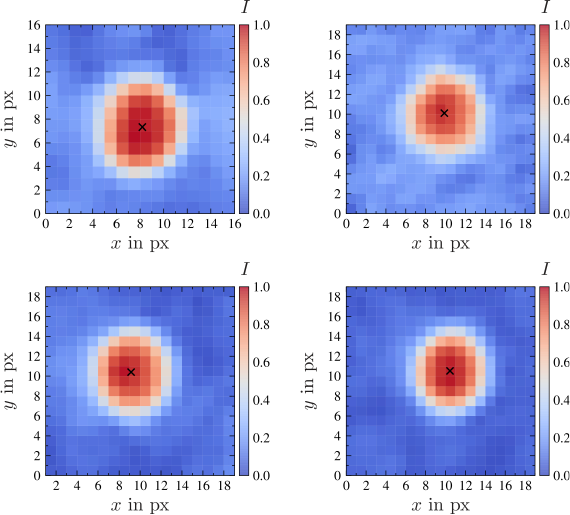}
        \caption{}
        \label{fig:refObj_exemplaryMarkers_IR}
    \end{subfigure}
    \caption{Reference object in IR image: a) full view of acquired IR image for camera model calibration and b) exemplary set of point markers after normalizing the image intensity to $I\in[0,1]$. Crosses indicate detected point centers using the intensity-based center of mass computation.}
    \label{fig:refObj_IR_expZamak}        
\end{figure}

In Sec.~\ref{sec:projection}, the pinhole camera model is described without accounting for image distortion, which must be justified for our experimental setup. Fig.~\ref{fig:refObj_fullIRview} visualizes the entire IR image of the reference object, which is used for the camera model calibration.
Apparently, the actual region of interest covered by the reference object (and subsequently the thin-walled tube specimen) represents only a small portion of the entire IR image. Note that moving the IR camera closer to the specimen would improve spatial resolution, but was not possible in the present experimental setup due to limitations on the available space for tripod placement around the tensile testing machine. As a result, an accurate calibration of lens distortion parameters in the small region of interest becomes very challenging. Moreover, the IR camera is equipped with a normal lens as specified in \ref{app:imagingSystems}. The reported field of view of $\ang{32.4}\times\ang{24.6}$ does not represent a wide-angle configuration. Hence, the IR camera was neither operated at a close range nor in a wide-angle configuration, justifying the neglect of lens distortion modeling in the present work.

\begin{figure}[htb]
    \centering
    \includegraphics[width=0.3\linewidth]{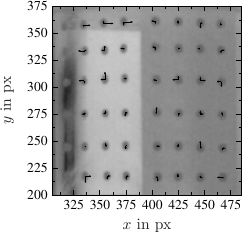}
    \caption{Directional leave-one-out cross-validation errors per marker in the IR image (scaled by factor $25$ for visualization)}
    \label{fig:refObj_loocv_intensityCom}
\end{figure}

To further quantify the projection accuracy without modeling lens distortion, we exploit leave-one-out cross-validation (LOOCV). An alternative, but less conservative, metric would be the reprojection error comparing the detected marker centers and predicted locations after calibrating the camera model. The projection matrix $\LM{M}$ is calibrated using all available points on the reference object except one, which is held out for validation. This procedure is repeated for every single point. Specifically, $42$ points are available in the present experimental setup on the surface of the reference object, omitting six points at the very left end, which are affected by reflections. Exemplary point markers in the IR image and their detected centers are shown in Fig.~\ref{fig:refObj_exemplaryMarkers_IR}, where the intensity-based center of mass computation is used to localize the centers, as explained in Sec.~\ref{sec:projection}. The LOOCV results are compiled in Tab.~\ref{tab:rejObj_pinhole_calibration} and align very well with the benchmark analysis on synthetic markers in \ref{app:marker_detection_IR}.
\begin{table}[htb]
    \centering
    \caption{Leave-one-out cross-validation results for IR marker detection (all quantities in \si{px})}
    \label{tab:rejObj_pinhole_calibration}
    \begin{tabular}{lcccccc}
        \toprule
        Method & RMSE $u^\Theta$ & RMSE $v^\Theta$ & Bias $u^\Theta$ & Bias $v^\Theta$ & Precision $u^\Theta$ & Precision $v^\Theta$ \\
        \midrule
        Canny + CoM & $0.2262$ & $0.1316$ & $0.0024$ & \cellcolor{lightgray}$-0.0019$ & $0.2289$ & $0.1331$ \\
        intensity + CoM & \cellcolor{lightgray}$0.1352$ & \cellcolor{lightgray}$0.1078$ & $0.0023$ & $-0.0026$ & \cellcolor{lightgray}$0.1369$ & \cellcolor{lightgray}$0.1091$ \\
        ellipse fit & $0.1540$ & $0.1251$ & \cellcolor{lightgray}$0.0016$ & $-0.0026$ & $0.1559$ & $0.1266$ \\
        2D-Gaussian fit & $0.1787$ & $0.1294$ & $0.0019$ & $-0.0029$ & $0.1808$ & $0.1310$ \\
        \bottomrule
    \end{tabular}
\end{table}
Note that $u^\Theta$ refers to the pixel coordinate coefficient in the horizontal, i.e., $x$-direction, and $v^\Theta$ similarly describes the vertical pixel coordinate coefficient. It turns out that the combination of Canny edge detection and computing the center of mass, which was put forth in \citep{hartmannmuellerlohsetroeger2023}, is the most inaccurate method, as it is highly affected by the low spatial resolution. While all methods achieve sub-pixel accuracy, the intensity-based center of mass computation achieves low error levels. Even more important, the near-zero mean bias indicates that no unmodeled systematic error is present, justifying the neglect of lens distortion modeling. In a related study, \citep{zengetal2023} reported reprojection errors of about \SI{0.2916}{\px}, which is outperformed here, even taking into account that the LOOCV errors are typically larger than pure reprojection errors. The directional LOOCV errors are visualized in Fig.~\ref{fig:refObj_loocv_intensityCom} for each accessible point marker in the IR image.
It becomes apparent that the errors are noise-like and do not exhibit any systematic error that could appear due to unmodeled nonlinear distortion effects. Hence, given the explanations on the IR camera specification and these quantitative results of LOOCV, the neglect of lens distortion modeling appears justified.

\subsection{Uncertainty Quantification}
\label{sec:exp_uq}
Before evaluating the results of the actual experiment, the associated uncertainties are quantified. The objective is two-fold: first, to provide the measurement uncertainties for the point coordinates $\LV{x}$ from stereocorrelation-based DIC and the pixel temperatures $\Theta$ from IRT, and second, to quantify the uncertainties in the postprocessed quantities at points on the curved surface, namely, maximum principal strains, point temperatures, in-plane temperature gradients, and temperature rates. The postprocessing, i.e., surface reconstruction and pixel temperature interpolation, is performed using the inverse multiquadric RBF
\begin{equation}
    \label{eq:IMQ}
    \hat{m}(\rho) = \frac{c_\mathrm{IMQ}}{\sqrt{c_\mathrm{IMQ}^2 + \rho^2}}.
\end{equation}
The shape factor $c_\mathrm{IMQ}$ is chosen as follows: $c_\mathrm{IMQ} = 0.9328$ (spatial) and $c_\mathrm{IMQ} = 1$ (temporal) are applied for the interpolation of the point coordinates, while $c_\mathrm{IMQ} = 2$ (spatial) and $c_\mathrm{IMQ} = 1$ (temporal) are used for the pixel temperature field. In \citep{hartmannmuellerlohsetroeger2021}, it was demonstrated that for the inverse multiquadric RBF the interpolation error is minimal when the shape factor is selected as $c_\mathrm{IMQ} = 2\Delta d/R_0$, where $\Delta d$ represents the point distance in the underlying data and $R_0$ the normalization coefficient introduced in Sec.~\ref{sec:interpRBF}. For our temporal data, image acquisition at $\SI{2}{\Hz}$ yields $\Delta d = \SI{0.5}{\s}$. With $\tilde{R}_0 = \SI{1}{\s}$, the shape factors for the temporal interpolation are set to $c_\mathrm{IMQ} = 1$. Similarly, the spatial shape factors mentioned above result from the point distance $\Delta d = \SI{0.4664}{\mm}$ in DIC (evaluated as the mean value on a projected plane for the curved surface coordinates) and $\Delta d = \SI{1}{\px}$ in IRT.

\begin{figure}[htb]
    \centering
    \begin{subfigure}[b]{0.45\linewidth}
        \centering
        \includegraphics[width=0.7\linewidth]{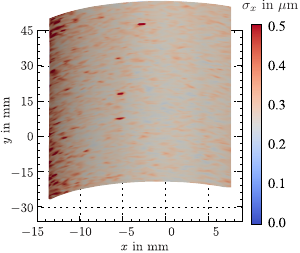}
        \caption{}
        \label{fig:uq_exp_x}        
    \end{subfigure}
    \hspace{0.04\textwidth}
    \begin{subfigure}[b]{0.45\linewidth}
        \centering
        \includegraphics[width=0.7\linewidth]{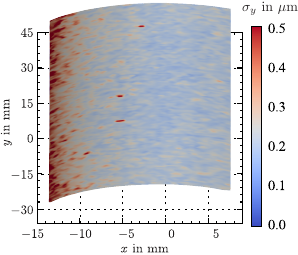}
        \caption{}
        \label{fig:uq_exp_y}
    \end{subfigure}

    \begin{subfigure}[b]{0.45\linewidth}
        \centering
        \includegraphics[width=0.7\linewidth]{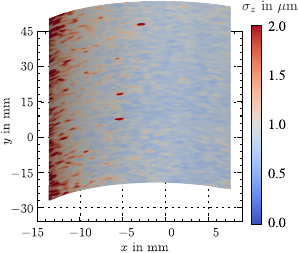}
        \caption{}
        \label{fig:uq_exp_z}        
    \end{subfigure}
    \hspace{0.04\textwidth}
    \begin{subfigure}[b]{0.45\linewidth}
        \centering
        \includegraphics[width=0.7\linewidth]{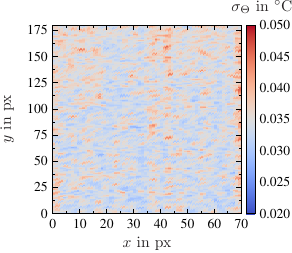}
        \caption{}
        \label{fig:uq_exp_Theta_px}
    \end{subfigure}
    \caption{Measurement uncertainties evaluated from $100$ images of the tube-like specimen in unloaded state: a) noise in $x$-coordinates, b) noise in $y$-coordinates, c) noise in $z$-coordinates, and d) pixel temperature noise}
    \label{fig:uq_exp}
\end{figure}

\paragraph{Experiment}
To determine the measurement uncertainties as noise-floor estimates, we acquire $100$ images of the thin-walled tube specimen without applying any loading, in line with established procedures in the literature, see exemplarily \citep{zaplaticetal2023}. This procedure allows us to compute the point coordinate uncertainties in stereocorrelation-based DIC for the present experimental setup and specimen. Similarly, the pixel temperature uncertainty in IRT is evaluated from $100$ images in the unloaded state, as carried out by \citep{zaplaticetal2023,cholewaetal2016}. The point coordinate uncertainties are visualized in Figs.~\ref{fig:uq_exp_x}-\ref{fig:uq_exp_z}. 
As is usually the case in stereocorrelation-based DIC, the uncertainty in the out-of-plane direction, here the $z$-direction, is larger than the uncertainties in the other two spatial directions. The spatial distribution of the point coordinate noise reveals that the left part of the specimen is primarily affected by noise, which could be caused by non-optimal illumination on this tube-like specimen in the present experimental setup. Nevertheless, the reported measurement uncertainties agree very well with reported literature values, see \citep{zaplaticetal2023}, for example. 

The pixel temperature uncertainty is shown in Fig.~\ref{fig:uq_exp_Theta_px} for the region of interest, i.e., the specimen surface. The observed noise level aligns very well with the IR camera specification of a temperature resolution below $\SI{0.05}{\K}$, see \ref{app:imagingSystems}. Consequently, performing IRT on the inhomogeneously varnished specimen surface does not significantly increase the observed pixel temperature fluctuations in the unloaded state. Furthermore, the IR camera's measurement accuracy of $\pm\SI{1.5}{\K}$ remains the dominant error source.

\begin{figure}[htb]
    \centering
    \begin{subfigure}[b]{0.45\linewidth}
        \centering
        \includegraphics[width=0.7\linewidth]{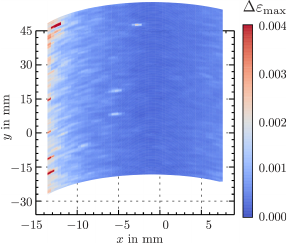}
        \caption{}
        \label{fig:uq_eval_strain}        
    \end{subfigure}
    \hspace{0.04\textwidth}
    \begin{subfigure}[b]{0.45\linewidth}
        \centering
        \includegraphics[width=0.7\linewidth]{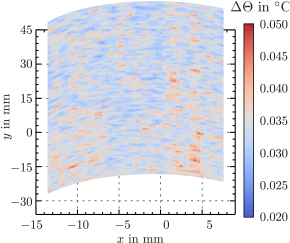}
        \caption{}
        \label{fig:uq_eval_temp}
    \end{subfigure}

    \begin{subfigure}[b]{0.45\linewidth}
        \centering
        \includegraphics[width=0.7\linewidth]{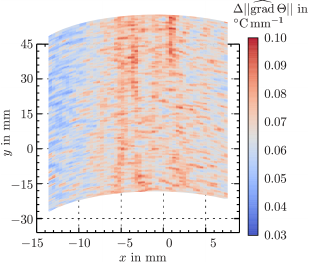}
        \caption{}
        \label{fig:uq_eval_tempGradNorm}        
    \end{subfigure}
    \hspace{0.04\textwidth}
    \begin{subfigure}[b]{0.45\linewidth}
        \centering
        \includegraphics[width=0.7\linewidth]{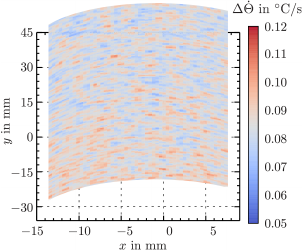}
        \caption{}
        \label{fig:uq_eval_tempRat}
    \end{subfigure}
    \caption{Uncertainties in postprocessed quantities: a) maximum principal strain, b) point temperature, c) norm of the temperature gradient, and d) temperature rate}
    \label{fig:uq_eval}
\end{figure}

\paragraph{Postprocessing}
Besides the measurement uncertainties, the question arises of how these uncertainties propagate through our postprocessing, i.e., surface reconstruction, temperature interpolation, strain and thermal analysis as shown in Fig.~\ref{fig:scheme}. Consequently, uncertainties in our quantities of interest must be assessed, specifically, maximum principal strains, point temperatures, in-plane temperature gradients, and temperature rates. From the $100$ images in the unloaded state, a noise-free reference is constructed by drawing on the mean values of both point coordinates and pixel temperatures. Considering this reference, the $100$ images in the unloaded state are considered as different realizations and evaluated using our postprocessing steps. As a result, the sample standard deviation yields the uncertainties as noise-floor estimates, see Fig.~\ref{fig:uq_eval}.
The uncertainty $\Delta{\varepsilon_\mathrm{max}}$ in the maximum principal strain is shown in Fig.~\ref{fig:uq_eval_strain}. It becomes evident that, as expected, the uncertainty remains very small, with a mean value of about $\num{7.e-04}$, reaching similar orders of magnitude as reported by \citep{zaplaticetal2023}. 

For our thermal analysis, the uncertainty in the point temperature in Fig.~\ref{fig:uq_eval_temp} is of particular interest. Compared to the pure pixel temperature noise in Fig.~\ref{fig:uq_exp_Theta_px}, it turns out that the projection of the pixel temperatures on the curved specimen surface does not increase the temperature uncertainty, reaching comparable values again as were reported by \citep{zaplaticetal2023} for the semi-hybrid multiview approach. Moreover, it must be remembered that the overall measurement accuracy of the employed IR camera is about $\pm\SI{1.5}{\K}$, which is therefore the main source of uncertainty in the temperature fields. As explained in Sec.~\ref{sec:tempGrad_tempVeloc}, we also evaluate the in-plane temperature gradient. For easier interpretation, the uncertainty in the norm of the in-plane temperature gradient is visualized in Fig.~\ref{fig:uq_eval_tempGradNorm}. The uncertainty is more pronounced at the center of the curved specimen surface, reaching about $\SI{0.1}{\celsius\per\mm}$. The uncertainty in the temperature rate $\dot{\Theta}$ is shown in Fig.~\ref{fig:uq_eval_tempRat}, with minimal values around $\SI{0.05}{\celsius\per\s}$ and maximum values of approximately $\SI{0.12}{\celsius\per\s}$.

To summarize the uncertainty quantification for the postprocessed full-field quantities, all quantities possess small uncertainties and can be accurately determined during the experimental tension-torsion test of the thin-walled tube specimen.

\paragraph{Impact of Projection}
Although the accuracy of the projected point temperatures has already been demonstrated above, the impact of the projection step should be quantified explicitly. Since only a single image is available per camera for the camera model calibration and the calibration is based on solving a homogeneous linear system, we employ a nonparametric bootstrap method to estimate the covariance matrix of the projection matrix coefficients. The covariance matrix is denoted as $\LM{C}_{\LV{m}}$, $\LM{C}_{\LV{m}}\in\realSetDim{12\times 12}$, using the column vector representation $\LV{m}$ of the projection matrix $\LM{M}$. The bootstrap approach assumes that each point pair of real-world coordinates from DIC and pixel coordinates from IRT is representative and independently and identically distributed. The procedure is selected since the camera model calibration using DLT, as explained in Sec.~\ref{sec:projection}, represents an \textit{errors-in-variables} setting since the uncertain point and pixel coordinates enter the system matrix $\LM{A}$, compare Eq.~\eqref{eq:coeffMat_homogenLinSys}. The bootstrap method inherently comprises measurement noise and point-configuration sensitivity. We exploit $1000$ bootstrap samples in total. For each sample, $42$ reference object points are drawn from the original dataset with replacement, representing the bootstrap dataset. Then, the DLT-based camera model calibration is carried out for each dataset. The sought covariance matrix $\LM{C}_{\LV{m}}$ results from the sample covariance.

\begin{figure}[htb]
    \centering
    \begin{subfigure}[b]{0.45\linewidth}
        \centering
        \includegraphics[width=0.7\linewidth]{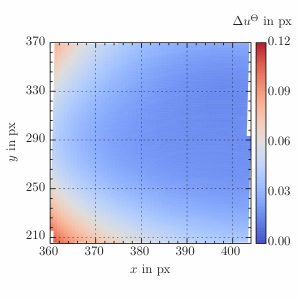}
        \caption{}
        \label{fig:uq_projMat_u}        
    \end{subfigure}
    \hspace{0.04\textwidth}
    \begin{subfigure}[b]{0.45\linewidth}
        \centering
        \includegraphics[width=0.7\linewidth]{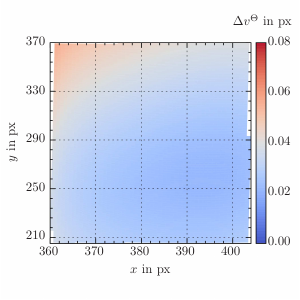}
        \caption{}
        \label{fig:uq_projMat_v}
    \end{subfigure}
    \caption{Uncertainty in pixel coordinates due to propagation of projection matrix coefficient uncertainties on the specimen surface: a) horizontal direction and b) vertical direction.}
    \label{fig:uq_projMat}
\end{figure}

The estimated covariance matrix $\LM{C}_{\LV{m}}$ allows for quantifying the pixel coordinate uncertainty in our present setup using Gaussian error propagation,
\begin{equation}
    \label{eq:GaussErrProp_dxTh}
    \LM{C}_{\LV{x}^\Theta} = \difn{\LV{x}^\Theta}{\LV{m}} \LM{C}_{\LV{m}} \left(\difn{\LV{x}^\Theta}{\LV{m}} \right)^\top.
\end{equation}
The covariance matrix $\LM{C}_{\LV{x}^\Theta}$, $\LM{C}_{\LV{x}^\Theta}\in\realSetDim{2\times 2}$, enables computing the pixel coordinate uncertainties $\Delta u^\Theta$ and $\Delta v^\Theta$ in horizontal and vertical direction, respectively. The derivative of the pixel coordinates with respect to the coefficients $\LV{m}$, i.e., $\D\LV{x}^\Theta/\D\LV{m}\in\realSetDim{2\times 12}$, is available analytically by deriving Eq.~\eqref{eq:perspectiveProjEq} and reads
\begin{equation}
    \label{eq:dxThdm}
    \difn{\LV{x}^\Theta}{\LV{m}} 
    =
    \frac{1}{\sigma(\LVq{x})}
    \begin{bmatrix}
	\LVq{x}^\top & \LV{0}^\top & -u^\Theta \LVq{x}^\top \\
	\LV{0}^\top & \LVq{x}^\top & -v^\Theta \LVq{x}^\top
	\end{bmatrix}.
\end{equation}
Note that the derivative must be evaluated at fixed projection matrix coefficients $\LV{m}$. The Gaussian error propagation is then performed for the mean values of the Cartesian coordinates $\LV{x}$, or their homogeneous counterparts $\LVq{x}$, determined from the $100$ images in the unloaded state. It should be mentioned that the pixel coordinates $u^\Theta$ and $v^\Theta$ can be formally substituted by the projection relation~\eqref{eq:perspectiveProjEq}. The resulting pixel coordinate uncertainties due to propagated projection matrix coefficient uncertainties are visualized in Fig.~\ref{fig:uq_projMat}, where the visualization of the projected Cartesian coordinates $\LV{x}$ into pixel space yields a slightly skewed visualization.
The results clearly demonstrate that the projection matrix $\LM{M}$ is accurately calibrated using the open book-shaped reference object. The uncertainties in horizontal pixel coordinates are larger than for the vertical direction. The upper-left corner of the specimen surface as the region of interest is the most sensitive domain in the present experiment, which was already the case for the LOOCV results in Fig.~\ref{fig:refObj_loocv_intensityCom}. Note that the evaluation is deliberately performed for the pixel coordinates $\LV{x}^\Theta$ instead of the projected point temperature to avoid the derivation of the temperature field in the unloaded state, which is not position-dependent apart from measurement noise. The small uncertainties in the pixel coordinates are important to ensure an accurate projection step when evaluating the interpolated pixel temperatures to assign point temperatures.

\subsection{Results}
\label{sec:exp_results}
\begin{figure}[ht]
    \centering
    \begin{subfigure}[b]{0.45\linewidth}
        \centering
        \includegraphics[width=0.7\linewidth]{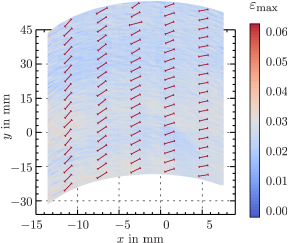}
        \caption{$t=\SI{4}{\s}$}
        \label{fig:zamak_priStrain_010}
    \end{subfigure}
    \hspace{0.04\textwidth}
    \begin{subfigure}[b]{0.45\linewidth}
        \centering
        \includegraphics[width=0.7\linewidth]{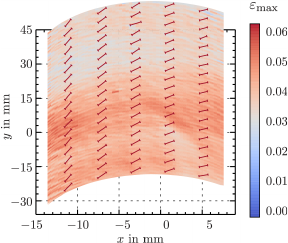}
        \caption{$t=\SI{8.5}{\s}$}
        \label{fig:zamak_priStrain_019}
    \end{subfigure}
    \caption{Maximum in-plane principal strain and principal strain directions on the curved surface of the tube-like specimen}
    \label{fig:zamak_priStrain}
\end{figure}

Similar to the uncertainty quantification above, the evaluation of the actual measurement is performed using the inverse multiquadric RBF~\eqref{eq:IMQ}. Of course, the selected shape factors remain unchanged. In addition to the thermal analysis, we perform RBF-based strain analysis as shown in Fig.~\ref{fig:scheme} and developed in \citep{hartmannmuellerlohsetroeger2021}. The force and torque response in Figs.~\ref{fig:zamak_force} and~\ref{fig:zamak_torque}, respectively, clearly indicate the introduced plastic deformations during the tension-torsion test. The full-field results are shown at two exemplary time points during loading ($t = \SI{4}{\s}$) and immediately after unloading ($t = \SI{8.5}{\s}$). For the strain analysis, the maximum in-plane principal strain $\varepsilon_\mathrm{max}$ is exemplarily visualized in Fig.~\ref{fig:zamak_priStrain}.
The small lines in Fig.~\ref{fig:zamak_priStrain} indicate the principal strain directions, which are typically challenging to compute in local interpolation approaches. Still, they are readily available here due to the global RBF-based ansatz \citep{hartmannmuellerlohsetroeger2021}. As is expected for a macroscopically homogeneous material in a tension-torsion test, the overall deformation is homogeneous in Fig.~\ref{fig:zamak_priStrain_010}. The principal strain directions also agree very well with the theoretical expectations for such loading conditions. These observations remain valid after unloading, compare Fig.~\ref{fig:zamak_priStrain_019}. 

\begin{figure}[ht]
    \centering
    \begin{subfigure}[b]{0.45\linewidth}
        \centering
        \includegraphics[width=0.7\linewidth]{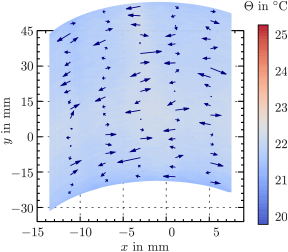}
        \caption{$t=\SI{4}{\s}$}
        \label{fig:zamak_temp-tempGrad_010}
    \end{subfigure}
    \hspace{0.04\textwidth}
    \begin{subfigure}[b]{0.45\linewidth}
        \centering
        \includegraphics[width=0.7\linewidth]{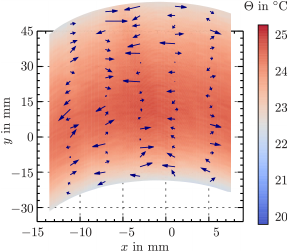}
        \caption{$t=\SI{8.5}{\s}$}
        \label{fig:zamak_temp-tempGrad_019}
    \end{subfigure}
    \caption{Temperature field and in-plane temperature gradient (scaled by a factor of 8) on the curved surface of the tube-like specimen}
    \label{fig:zamak_temp-tempGrad}
\end{figure}

\begin{figure}[ht]
    \centering
    \begin{subfigure}[b]{0.45\linewidth}
        \centering
        \includegraphics[width=0.7\linewidth]{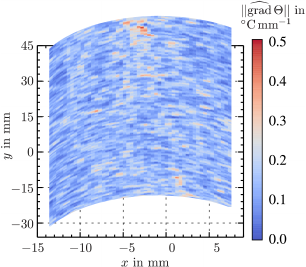}
        \caption{$t=\SI{4}{\s}$}
        \label{fig:zamak_tempGrad_010}
    \end{subfigure}
    \hspace{0.04\textwidth}
    \begin{subfigure}[b]{0.45\linewidth}
        \centering
        \includegraphics[width=0.7\linewidth]{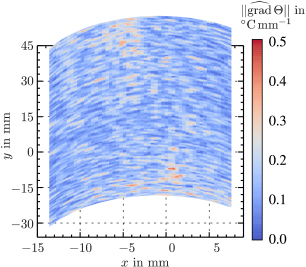}
        \caption{$t=\SI{8.5}{\s}$}
        \label{fig:zamak_tempGrad_019}
    \end{subfigure}
    \caption{Norm of in-plane temperature gradient on the curved surface of the tube-like specimen}
    \label{fig:zamak_tempGrad}
\end{figure}

In compliance with the focus of this work, the temperature field and the associated in-plane temperature gradients are shown in Fig.~\ref{fig:zamak_temp-tempGrad}. 
A homogeneous temperature field is observed during both loading and unloading in Figs.~\ref{fig:zamak_temp-tempGrad_010} and~\ref{fig:zamak_temp-tempGrad_019}, respectively. Considering the expected homogeneous deformation in tension-torsion of a thin-walled tube, this observation is very reasonable. In Fig.~\ref{fig:zamak_temp-tempGrad_019}, small regions with slightly reduced temperatures are observable at the top and bottom in the $y$-direction. These reductions could be attributed to heat conduction to the clampings or edge effects of the RBF-based interpolation, which suffers from reduced accuracy towards the boundary of the region of interest \citep{hartmannmuellerlohsetroeger2021}. Moreover, the norm of the in-plane temperature gradients is depicted in Fig.~\ref{fig:zamak_tempGrad}.
As a result of the comparatively homogeneous temperature distribution, the norm of the temperature gradient remains relatively small. Compared to the estimated uncertainty in Fig.~\ref{fig:uq_eval_tempGradNorm}, the determined temperature gradients are quite uncertain, which is caused by the homogeneous temperature distribution and not an artifact of the RBF-based surface interpolation and thermal analysis.

\begin{figure}[ht]
    \centering
    \includegraphics[width=0.28\linewidth]{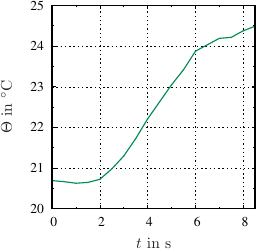}
    \caption{Temperature evolution of a point centered in the curved surface of the tube-like specimen}
    \label{fig:zamak_temp-time}
\end{figure}

The temporal evolution of the temperature is shown in Fig.~\ref{fig:zamak_temp-time} for a point centered on the observed curved surface.
At the beginning of the loading, a very small temperature decrease can be observed, which could be attributed to the Gough-Joule effect or noise in the acquired pixel temperatures. Subsequently, a notable temperature increase is observed due to plastic loading. Since only the surface can be observed, the temperature increases further even after the loading has finished at $t = \SI{5}{\s}$. 

\begin{figure}[ht]
    \centering
    \begin{subfigure}[b]{0.45\linewidth}
        \centering
        \includegraphics[width=0.7\linewidth]{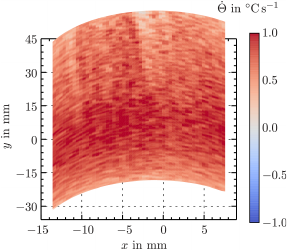}
        \caption{$t=\SI{4}{\s}$}
        \label{fig:zamak_tempRate_010}
    \end{subfigure}
    \hspace{0.04\textwidth}
    \begin{subfigure}[b]{0.45\linewidth}
        \centering
        \includegraphics[width=0.7\linewidth]{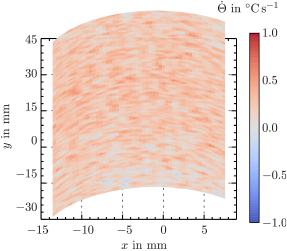}
        \caption{$t=\SI{8.5}{\s}$}
        \label{fig:zamak_tempRate_019}
    \end{subfigure}
    \caption{Temperature rate on the curved surface of the tube-like specimen}
    \label{fig:zamak_tempRate}
\end{figure}

The temperature rate is illustrated in Fig.~\ref{fig:zamak_tempRate}.
The temperature rate during loading in Fig.~\ref{fig:zamak_tempRate_010} is comparatively homogeneous, reaching a temperature rate of about $\SI{1}{\celsius\per\second}$, see Fig.~\ref{fig:zamak_temp-time} as well. As expected, the temperature rate decreases and approaches zero immediately after unloading, see Fig.~\ref{fig:zamak_tempRate_019}. In relation to the temperature rate uncertainty estimate in Fig.~\ref{fig:uq_eval_tempRat}, the temperature rate during unloading is close to the noise level, which is reasonable since no further plastic dissipation happens during unloading.

The present example of a tube-like specimen subjected to tension-torsion load demonstrates the applicability of both projection-based coupling and RBF-based global interpolation concept in experimental mechanics. While many studies evaluate at most the displacement or strain, and the temperature field, we determine a comprehensive set of full-field data, which is valuable for subsequent steps such as model calibration or validation.

\section{Conclusions and Outlook}
\label{sec:conclusions}
We address the increasing demand for thermo-mechanical data for constitutive model calibration, discovery, and validation by proposing a straightforward yet accurate projection-based coupling of stereocorrelation-based digital image correlation and infrared thermography. Our projection-based approach enables us to couple point coordinates from digital image correlation with pixel temperatures from infrared thermography to express the temperature field in the same Lagrangian frame as the mechanical fields. To perform the projection, the pinhole camera model must be calibrated first using a single image of a three-dimensional calibration object. Subsequently, the individually calibrated imaging systems are employed for image acquisition and registration. Our approach is entirely external, i.e., coupling the data rather than the imaging systems, applicable to flat and curved surfaces, and essentially constitutes an additional postprocessing step, which can be readily integrated into existing experimental protocols. 

Apart from the projection-based coupling of mechanical and thermal full-field data, we have extended the radial basis function-based evaluation of full-field data to spatial and temporal interpolation. For the temperature field, this approach enables the evaluation of in-plane temperature gradients and temperature rates on flat and curved surfaces. By doing so, we extend thermo-mechanical datasets beyond displacement, strain, and temperature fields and provide additional full-field data for the aforementioned tasks of calibration and validation. In a certain sense, we do not measure more, but take the most out of the acquired full-field data.

As future work, the experimental design will be optimized to obtain heterogeneous fields in thermo-mechanical tests following the suggestion by \citep{lambrughithuilliercoppieters2025}. Regarding the interpolation concept, smoothing techniques by means of regression rather than interpolation are of particular interest, especially in the temporal domain.

\appendix

\section{Marker Detection in IR Image}
\label{app:marker_detection_IR}
The accurate calibration of the pinhole camera model, i.e., determining the projection matrix $\LM{M}$ in Eq.~\eqref{eq:perspectiveProjEq}, crucially relies on the accuracy of the point coordinates $\LV{x}$ and the pixel coordinates $\LV{x}^\Theta$. The former are accurately determined using stereocorrelation-based DIC and the point markers, see Fig.~\ref{fig:refObj}. However, computing the pixel coordinates $\LV{x}^\Theta$ in the IR image is more involved due to the significantly lower spatial resolution compared to DIC. 

Localizing the center of circular or elliptical objects is a long-standing challenge when using optical methods, see, for example, \citep{vanassenetal2002,jobbagyfurnee1994}, and remains an active area of research, we refer to \citep{hanetal2026,cietal2026} for recent studies and an overview of the current state of the art. Center extraction methods can be distinguished into two categories: i) edge-based techniques that extract edge contours and subsequently fit geometric features, and ii) intensity-based methods that infer the geometric center by relying on (gray-scale) intensity distributions \citep{cietal2026}. Due to the importance of the pixel coordinates $\LV{x}^\Theta$, we assess four different methods for localizing the center of our point markers in the IR image, compare Fig.~\ref{fig:refObj}:
\begin{itemize}
    \item Canny edge detection + center of mass computation
    \item Intensity-based weighted center of mass computation
    \item Canny edge detection + least-squares fit of an ellipse
    \item Least-squares-based fit of a 2D-Gaussian
\end{itemize}
The first two methods are inspired by \citep{gohaneetal2016}. Canny edge detection yields a mask of pixels associated with the marker edge and allows for straightforward computation of the center of mass. This method was already used in earlier work \citep{hartmannmuellerlohsetroeger2023} but is inferior in a low-resolution setting, which holds likewise for other edge-based approaches \citep{cietal2026}. As a remedy, intensity-based weighted center of mass computation is more robust since the image intensity after normalization is taken into account. Instead of computing the center of mass after Canny edge detection, an ellipse can be determined by least-squares fitting. The fourth center localization method investigated here is fitting a $2$D-Gaussian to the normalized image intensity values.

\begin{figure}[ht]
    \centering
    \begin{subfigure}[b]{0.35\linewidth}
        \centering
        \includegraphics[width=0.7\linewidth]{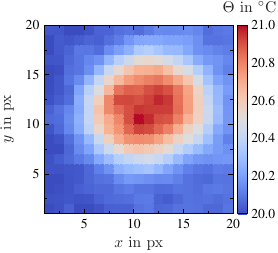}
        \caption{}
        \label{fig:synMarker_cold}
    \end{subfigure}
    \hspace{0.1\textwidth}
    \begin{subfigure}[b]{0.35\linewidth}
        \centering
        \includegraphics[width=0.7\linewidth]{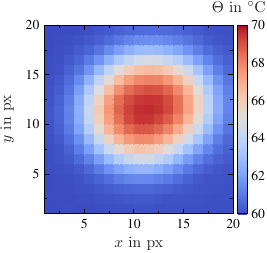}
        \caption{}
        \label{fig:synMarker_warm}
    \end{subfigure}
    \caption{Exemplary synthetic markers representing a) an unheated reference object and b) a heated reference object}
    \label{fig:synthetic_markers}
\end{figure}

For comparability, the four methods are evaluated using a synthetic benchmark example of an elliptical marker. A single marker is synthetically generated on a $\num{20}\times\SI{20}{\px}$ region of interest. The elliptical shape with semi-axes of $\SI{5}{\px}$ and $\SI{6}{\px}$ and a rotation of $\ang{15}$ against the horizontal axis is deliberately chosen as a challenging example compared to the observed markers in our experimental setup in Fig.~\ref{fig:refObj_exemplaryMarkers_IR}. The synthetic image is superimposed with a Gaussian blur with standard deviation $\SI{2}{\px}$, additive Gaussian noise of $\SI{0.05}{\celsius}$, and a linear background temperature gradient in the horizontal direction with $\SI{0.005}{\celsius\per\px}$ to represent a realistic scenario. Moreover, we evaluate two settings: an unheated reference object (background $\SI{20}{\celsius}$ and marker $\SI{21}{\celsius}$) and a heated reference object (background $\SI{60}{\celsius}$ and marker $\SI{70}{\celsius}$). Exemplary synthetic markers are visualized in Fig.~\ref{fig:synthetic_markers} for both cases.
The marker center is shifted by $10$ sub-pixel offsets in each direction. Moreover, $50$ realizations of additive Gaussian noise are used, i.e., in total $5000$ synthetic markers are generated with known center coordinates. The generated images are preprocessed by removing the background with a least-squares fitting of a plane, which is linear in both directions, and normalizing the image data to an intensity $I \in [0,1]$.

\begin{figure}[ht]
    \centering
    \begin{subfigure}[b]{0.4\linewidth}
        \centering
        \includegraphics[width=0.7\linewidth]{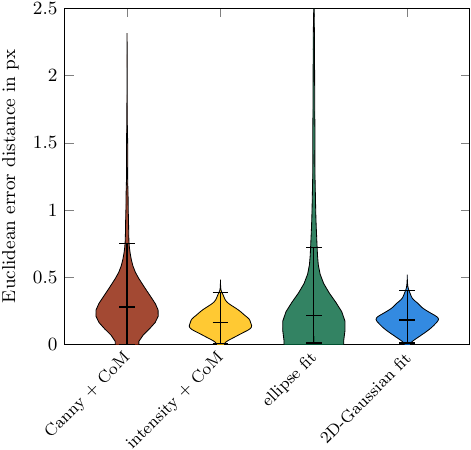}
        \caption{}
        \label{fig:violin_benchmark_cold}
    \end{subfigure}
    \hspace{0.1\textwidth}
    \begin{subfigure}[b]{0.4\linewidth}
        \centering
        \includegraphics[width=0.7\linewidth]{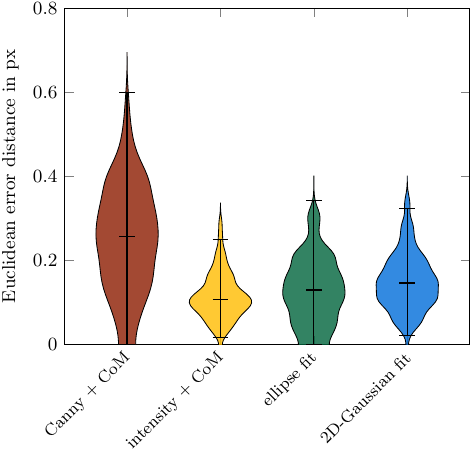}
        \caption{}
        \label{fig:violin_benchmark_warm}
    \end{subfigure}
    \caption{Distributions of Euclidean error distance for all four methods for a) the unheated and b) the heated reference object. Whiskers denote the most extreme data points within 1.5 times the interquartile range from the first and third quartiles (Tukey's rule).}
    \label{fig:violins_benchmark}
\end{figure}

To assess the performance of the four different markers for center localization in the IR image, we evaluate the root mean square error (RMSE) against the known center location to determine the typical error magnitude. For a comprehensive evaluation, directional bias (error mean) and precision (error standard deviation) are reported as well. The results are compiled in Tab.~\ref{tab:benchmark_marker} for the unheated and heated reference object.
\begin{table}[htb]
    \centering
    \caption{Benchmark analysis results for IR marker detection (all quantities in \si{px})}
    \label{tab:benchmark_marker}
    \begin{tabular}{lcccccc}
        \toprule
        Method & RMSE $u$ & RMSE $v$ & Bias $u$ & Bias $v$ & Precision $u$ & Precision $v$ \\
        \midrule
        \multicolumn{7}{l}{unheated reference object} \\
        \midrule
        Canny + CoM & $0.2992$ & $0.2984$ & $-0.0438$ & \cellcolor{lightgray}$-0.0699$ & $0.2960$ & $0.2900$ \\
        intensity + CoM & \cellcolor{lightgray}$0.1034$ & \cellcolor{lightgray}$0.1532$ & \cellcolor{lightgray}$-0.0412$ & $-0.1276$ & $0.0948$ & $0.0849$ \\
        ellipse fit & $0.4211$ & $0.4047$ & $-0.0708$ & $-0.0878$ & $0.4152$ & $0.3951$ \\
        $2$D-Gaussian fit & $0.1151$ & $0.1628$ & $-0.0680$ & $-0.1407$ & \cellcolor{lightgray}$0.0929$ & \cellcolor{lightgray}$0.0819$ \\
        \midrule
        \multicolumn{7}{l}{heated reference object} \\
        \midrule
        Canny + CoM & $0.2303$ & $0.1744$ & $0.0107$ & \cellcolor{lightgray}$0.0156$ & $0.2300$ & $0.1737$ \\
        intensity + CoM & \cellcolor{lightgray}$0.0851$ & $0.0928$ & $-0.0361$ & $-0.0697$ & \cellcolor{lightgray}$0.0770$ & \cellcolor{lightgray}$0.0613$ \\
        ellipse fit & $0.1288$ & \cellcolor{lightgray}$0.0845$ & \cellcolor{lightgray}$-0.0013$ & $-0.0187$ & $0.1288$ & $0.0824$ \\
        $2$D-Gaussian fit & $0.1056$ & $0.1216$ & $-0.0636$ & $-0.1015$ & $0.0843$ & $0.0670$ \\
        \bottomrule
    \end{tabular}
\end{table}
The specific error distributions for the Euclidean error distance as a combined isotropic error magnitude are visualized in Fig.~\ref{fig:violins_benchmark}.

The results in Tab.~\ref{tab:benchmark_marker} and Fig.~\ref{fig:violins_benchmark} clearly demonstrate that a heated reference object leads to lower localization errors, which is in line with earlier studies on coupling DIC and IRT \citep{cholewaetal2016,zengetal2023}. Nevertheless, the unheated reference object yields subpixel accuracy, even with a very small difference between marker and background. Therefore, we employ an unheated reference object in the present study. As shown in Fig.~\ref{fig:violins_benchmark}, the four different methods exhibit a strong variation in their accuracy. As noted by \citep{cietal2026}, edge-based methods, here Canny + CoM and the ellipse fit, suffer from low spatial resolution. In contrast, the intensity-based center of mass computation and fitting a $2$D-Gaussian yield accurate results, independent of whether a heated or unheated reference object is employed. It turns out that the intensity-based center of mass computation is the most suitable method for the present example of localizing the center of a marker in an IR image. Hence, we rely on the intensity-based approach to determine the pixel coordinates $\LV{x}^\Theta$ of the markers on the reference object in the IR image.

\section{Continuous Description of Curved Surfaces}
\label{app:curvedKinematics}
\begin{figure}[ht]
    \centering
    \includegraphics[width=0.75\textwidth]{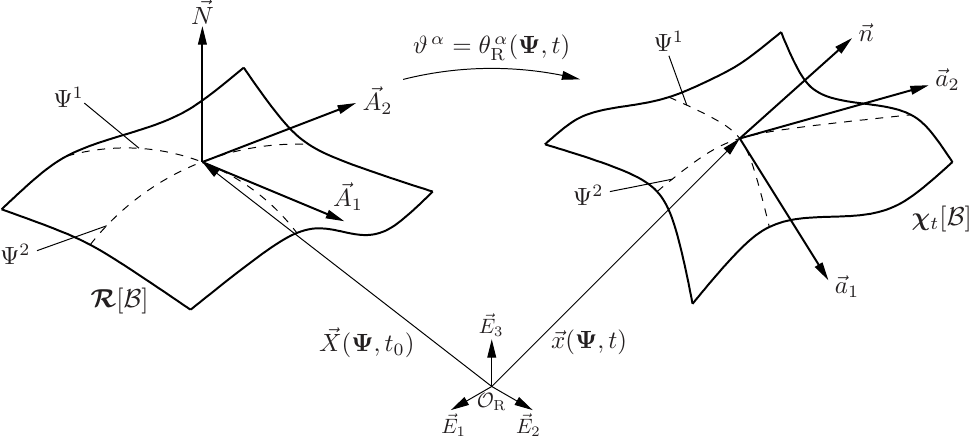}
    \caption{Kinematics of curved surfaces}
    \label{fig:kinematicsCurvedSurf}
\end{figure}

A continuous description of curved surfaces is fundamental for computing strains or, as in this work, assigning temperature values to material points and computing spatial temperature gradients and temperature rates within these surfaces. The general description and kinematics of curved surfaces are illustrated in Fig.~\ref{fig:kinematicsCurvedSurf}.
Using convective coordinates, a curvilinear surface is parametrized by the surface parameters $\bm{\Psi} = \lbrace \Psi^1,\Psi^2 \rbrace^\top$ in both reference configuration $\bm{\mathcal{R}}[\mathcal{B}]$ and current configuration $\bm{\chi}_t[\mathcal{B}]$ of the material body $\mathcal{B}$. Note that the reference configuration is assumed to coincide with the initial configuration $\bm{\chi}_{t_0}[\mathcal{B}]$ at $t_0$. The geometrical vectors $\vec{X}(\bm{\Psi},t_0)$ and $\vec{x}(\bm{\Psi},t)$ are expressed by means of the surface parameters $\bm{\Psi}$. The motion is defined by $\theta_\mathrm{R}^\alpha(\bm{\Psi},t)$. The tangent vectors at a surface point are given as
\begin{equation}
    \label{eq:tangentVectors}
    \vec{A}_\alpha = \dif{\vec{X}(\Psi^\alpha,t_0)}{\Psi^\alpha},
    \qquad
    \vec{a}_\alpha = \dif{\vec{x}(\Psi^\alpha,t)}{\Psi^\alpha}.
\end{equation}
Here, we proceed according to the common convention and count Greek letters from 1 to 2. Computing tangent vectors requires a continuous surface description in terms of surface parameters, underscoring the need to interpolate the discrete pointwise DIC data to reconstruct the specimen's surface. The covariant metric coefficients in both configurations are defined by the scalar products, $A_{\alpha\beta} = \vec{A}_\alpha\cdot\vec{A}_\beta$ and $a_{\alpha\beta} = \vec{a}_\alpha\cdot\vec{a}_\beta$, respectively. These coefficients allow the computation of the gradient vectors $\vec{A}^\alpha$ and $\vec{a}^\alpha$ by means of the contravariant metric coefficients $\left[A^{\alpha\beta}\right] = \left[A_{\alpha\beta}\right]^{-1}$ and $\left[a^{\alpha\beta}\right] = \left[a_{\alpha\beta}\right]^{-1}$,
\begin{equation}
    \label{eq:gradientVectors}
    \vec{A}^\alpha = A^{\alpha\beta} \vec{A}_\beta,
    \qquad
    \vec{a}^\alpha = a^{\alpha\beta} \vec{a}_\beta.
\end{equation}
During strain analysis, the in-plane deformation gradient $\Th{F} = \vec{a}_\alpha \otimes \vec{A}^\alpha$ is an essential quantity. See \citep{hartmannmuellerlohsetroeger2021} for further details on strain analysis in curved surfaces. In this work, the gradient vectors are required for the computation of the in-plane temperature gradients~\eqref{eq:tempGrad}.

\section{Derivative of Projection Equation}
\label{app:derivativeProjection}
Both the spatial in-plane temperature gradient~\eqref{eq:tempGrad} and the temperature rate~\eqref{eq:tempVeloc} require the derivative of the pixel coordinates $\LV{x}^\Theta$ in the IR data with respect to the real-world coordinates $\LV{x}$, which were obtained from stereocorrelation-based DIC. The result is stated in Eq.~\eqref{eq:derivProjection} with its derivation explained below. 

Since the projection equation~\eqref{eq:perspectiveProjEq} is formulated by means of homogeneous coordinates $\LVq{x}^\Theta$ and $\LVq{x}$, respectively, we seek a formulation using directly the pixel and real-world coordinates $\LV{x}^\Theta$ and $\LV{x}$ instead. For that, the projection matrix is partitioned,
\begin{equation}
    \label{eq:reformProjMat}
    \renewcommand{\arraystretch}{1.25}
    \LM{M} = \begin{bmatrix}
        \LVt{m}_1^\top & m_{14} \\
        \LVt{m}_2^\top & m_{24} \\
        \LVt{m}_3^\top & m_{34}
    \end{bmatrix}
\end{equation}
with the column vectors $\LVt{m}_k\in\mathbb{R}^3$, $k=1,2,3$. Then, the projection equation~\eqref{eq:perspectiveProjEq} can be reformulated,
\begin{equation}
    \label{eq:reformProjEq}
    \renewcommand{\arraystretch}{1.25}
    \LV{x}^\Theta = \frac{1}{\sigma(\LV{x})}
    \left(
    \begin{bmatrix}
    \LVt{m}_1^\top \\ \LVt{m}_2^\top    
    \end{bmatrix}
    \LV{x}
    +
    \begin{Bmatrix}
        m_{14} \\ m_{24}
    \end{Bmatrix}
    \right)
    =
    \frac{1}{\LVt{m}_3^\top\LV{x}+m_{34}}
    \left(
    \begin{bmatrix}
    \LVt{m}_1^\top \\ \LVt{m}_2^\top    
    \end{bmatrix}
    \LV{x}
    +
    \begin{Bmatrix}
        m_{14} \\ m_{24}
    \end{Bmatrix}
    \right).
\end{equation}
Note that for the scale factor $\sigma(\LV{x}) = \LV{m}_3^\top\LVq{x} = \LVt{m}_3^\top\LV{x} + m_{34}$ holds. Concisely, the derivative~\eqref{eq:derivProjection} follows from relation~\eqref{eq:reformProjEq} using the quotient rule. For example, we obtain for the first pixel coordinate, $x_1^\Theta = u^\Theta$,
\begin{equation}
    \label{eq:detailedDerivPixCoord}
    \difn{u^\Theta}{\LV{x}} = \frac{\LVt{m}_1^\top \sigma(\LV{x}) - (\LVt{m}_1^\top\LV{x}+m_{14})\LVt{m}_3^\top}{\left(\sigma(\LV{x})\right)^2} \\
    = \frac{1}{\sigma(\LV{x})}\left(\LVt{m}_1^\top - u^\Theta\LVt{m}_3^\top\right).
\end{equation}
The result for the second pixel coordinate $v^\Theta$ is calculated similarly.

\section{Derivatives using Radial Basis Function-Based Interpolation}
\label{app:derivativesRBFInterp}
To compute the temperature gradient and temperature rate on a, in general, curved surface, several derivatives are required, necessitating continuous surface interpolation. The following derivatives occur in the chosen RBF-based interpolation. The derivative $\D\hat{m}/\D\rho$ is available in an analytical form for the specific RBF, see, for example, \citep{muellerlohsetroegerhartmann2023a}.

In Eq.~\eqref{eq:dThetadxTh}, the normalized distance $\check{\rho}$ must be derived with respect to the coordinates of the pixel coordinate vector $\LV{x}^\Theta$,
\begin{equation}
    \label{eq:drho_dxTh}
    \difn{\check{\rho}}{\LV{x}^\Theta} = \left[ \dif{\check{\rho}}{u^\Theta} \dif{\check{\rho}}{v^\Theta} \right],
\end{equation}
where the individual derivatives read
\begin{equation}
    \label{eq:drho_duvTh}
    \dif{\check{\rho}}{u^\Theta} = \frac{u^\Theta - u^\Theta_k}{\check{R}_0\vert\vert \LV{x}^\Theta - \LV{x}^\Theta_k \vert\vert} 
    \quad
    \text{and}
    \quad
    \difn{\check{\rho}}{v^\Theta} = \frac{v^\Theta - v^\Theta_k}{\check{R}_0\vert\vert \LV{x}^\Theta - \LV{x}^\Theta_k \vert\vert}.
\end{equation}

The local derivative~\eqref{eq:dThetadt} of the temperature field with respect to time is required to compute the temperature rate~\eqref{eq:tempVeloc}. Therein, the specific ansatz for interpolation in the temporal domain has to be derived,
\begin{align}
    \label{eq:daTheta_dt}
    \difn{a_k^\Theta}{t} &= \sum_{i=1}^{n_\mathrm{RBF}^\mathrm{t}} \difn{\hat{m}}{\rho}\dif{\tilde{\rho}}{t} \hat{a}_i^\mathrm{t} + \sum_{j=1}^{n_\mathrm{Mon}^\mathrm{t}} \difn{\tilde{n}_j}{t} \check{a}_j^\mathrm{t}, \\
    \label{eq:dbTheta_dt}
    \difn{b_l^\Theta}{t} &= \sum_{i=1}^{n_\mathrm{RBF}^\mathrm{t}} \difn{\hat{m}}{\rho}\dif{\tilde{\rho}}{t} \hat{b}_i^\mathrm{t} + \sum_{j=1}^{n_\mathrm{Mon}^\mathrm{t}} \difn{\tilde{n}_j}{t} \check{b}_j^\mathrm{t}.
\end{align}

Furthermore, the continuous interpolation of the curved surface must be derived with respect to the surface parameters in Eq.~\eqref{eq:dxdPsi}. This term contains the derivative of the normalized distance $\hat{\rho}$ with respect to the surface parameters $\Psi^\alpha$,
\begin{equation}
    \label{eq:drho_dPsi}
    \dif{\hat{\rho}}{\Psi^\alpha} = \frac{\Psi^\alpha - \Psi^\alpha_k}{R_0\vert\vert \bm{\Psi}-\hat{\bm{\Psi}}_k \vert\vert}.
\end{equation}

Finally, the temporal derivative \eqref{eq:dxdt} of the surface interpolation, which accounts for the temporal evolution of the surface, is required and comprises the derivative of the time-dependent coefficients, 
\begin{align}
    \label{eq:dAs_dt}
    \difn{A_{kj}^\mathrm{s}}{t} &= \sum_{i=1}^{n_\mathrm{RBF}^\mathrm{t}} \difn{\hat{m}}{{\rho}} \dif{\tilde{\rho}}{t} a_i^\mathrm{t} + \sum_{j=1}^{n_\mathrm{Mon}^\mathrm{t}} \difn{\tilde{n}_j}{t} \tilde{a}_j^\mathrm{t}, \\
    \label{eq:dBs_dt}
    \difn{B_{lj}^\mathrm{s}}{t} &= \sum_{i=1}^{n_\mathrm{RBF}^\mathrm{t}} \difn{\hat{m}}{{\rho}} \dif{\tilde{\rho}}{t} b_i^\mathrm{t} + \sum_{j=1}^{n_\mathrm{Mon}^\mathrm{t}} \difn{\tilde{n}_j}{t} \tilde{b}_j^\mathrm{t}.
\end{align}

Note that the derivatives of the monomials \eqref{eq:monomials_space}, \eqref{eq:monomials_time}, and \eqref{eq:monomials_Theta} are not explicitly restated here because they are considered trivial for the present application of at most linear monomials.

\section{Imaging Systems and Settings for Image Registration}
\label{app:imagingSystems}
During both experiments, the same camera systems are utilized. Specifically, the IR camera (VarioCAM HD, InfraTec GmbH, Dresden, Germany) operated within the spectral range $\num{7.5}\ldots\SI{14}{\micro\m}$ and was equipped with a normal lens (focal length $\SI{30}{\mm}$), yielding a field of view $\ang{32.4}\times\ang{24.6}$ and an instantaneous field of view $\SI{0.57}{\milli\radian}$. The detector had a format of $\num{1024}\times\SI{768}{\px}$. The measuring distance is negligible because the transmittance of air as a surrounding medium is very close to one in the employed long-wave IR window. The temperature resolution of the IR camera is below $\SI{0.05}{\K}$ with a measurement accuracy of $\pm\SI{1.5}{\K}$.

The stereocorrelation-based DIC system (ARAMIS 12M, Carl Zeiss GOM Metrology GmbH, Brunswick, Germany) comprised two visible light cameras ($\num{4096}\times\SI{3000}{\px}$, $\SI{50}{\mm}$ lenses). The illumination was performed using an ARAMIS light projector (Carl Zeiss GOM Metrology GmbH, Brunswick, Germany) with blue-light technology. The images were acquired at $\SI{2}{\Hz}$, of course, employing triggered image capture to synchronize both camera systems. The subset-based image correlation is performed with a facet size of \SI{19}{\px} and a point distance of \SI{15}{\px} between the centers of adjacent facets, resulting in partially overlapping facets.

\bibliographystyle{ieeetr}
\bibliography{literature}

\end{document}